\documentclass{article}

\usepackage[utf8]{inputenc} 
\usepackage[T1]{fontenc}    
\usepackage{hyperref}       
\usepackage{url}            
\usepackage{booktabs}       
\usepackage{amsfonts}       
\usepackage{nicefrac}       
\usepackage{microtype}      
\usepackage{xcolor}   

\usepackage{float}
\restylefloat{table}

\usepackage{amssymb}
\usepackage{amsmath}
\usepackage{amsthm}
\theoremstyle{plain}
\usepackage{bbm}
\usepackage{mathtools}
\usepackage{bm}
\usepackage{csvsimple}
\usepackage{dirtytalk}
\usepackage{mdframed}
\usepackage{ulem}
\usepackage{tikz}
\usetikzlibrary{arrows,backgrounds}
\usepgflibrary{shapes.multipart}

\usepackage{graphicx}
\usepackage{mdframed}
\usepackage{stmaryrd} 
\usepackage{pifont}
\usepackage{cancel}

\newcommand{\Portable}{{\texttt{ NVIDIA RTX 2080 MaxQ}}}
\newcommand{\VM}{{\texttt{Cloud Computing}}}

\newcommand{\UCIa}{{\texttt{Concrete Slump Test$-$3}}}
\newcommand{\UCIb}{{\texttt{QSAR aquatic toxicity}}}
\newcommand{\UCIc}{{\texttt{Seoul Bike Sharing Demand}}}
\newcommand{\UCId}{{\texttt{Beijing PM2.5 Data}}}

\newcommand{\Rdeux}{{\texttt{R}^2}}
\newcommand{\OLS}{{\texttt{OLS}}}
\newcommand{\SVM}{{\texttt{SVM}}}
\newcommand{\DL}{{\texttt{DL}}}
\newcommand{\GBDT}{{\texttt{GBDT}}}
\newcommand{\DNDT}{{\texttt{DNDT}}}
\newcommand{\AM}{{\texttt{AM}}}
\newcommand{\NLP}{{\texttt{NLP}}}
\newcommand{\TabNN}{{\texttt{TabNN}}}
\newcommand{\TabNet}{{\texttt{TabNet}}}
\newcommand{\DNN}{{\texttt{DNN}}}
\newcommand{\NODE}{{\texttt{NODE}}}
\newcommand{\Catboost}{{\texttt{CatBoost}}}
\newcommand{\XGBoost}{{\texttt{XGBoost}}}
\newcommand{\lightgbm}{{\texttt{LightGBM}}}
\newcommand{\RLN}{{\texttt{RLN}}}
\newcommand{\CNN}{{\texttt{CNN}}}
\newcommand{\NTK}{{\texttt{NTK}}}
\newcommand{\NetDNF}{{\texttt{Net-DNF}}}
\newcommand{\SNN}{{\texttt{SNN}}}
\newcommand{\RBF}{{\texttt{RBF}}}

\newcommand{\SELU}{{\texttt{SELU}}}
\newcommand{\NAN}{{\texttt{NAN}}}

\newcommand{\Pa}{{\texttt{P90}}}
\newcommand{\Pb}{{\texttt{P95}}}
\newcommand{\Pc}{{\texttt{P98}}}
\newcommand{\PMA}{{\texttt{PMA}}}
\newcommand{\FR}{{\texttt{F. Rank}}}
\newcommand{\AUC}{{\texttt{AUC}}}
\newcommand{\Acc}{{\texttt{Acc.}}}

\newcommand{\RF}{{\texttt{RF}}}
\newcommand{\XRF}{{\texttt{XRF}}}
\newcommand{\MLP}{{\texttt{MLP}}}
\newcommand{\XGB}{{\texttt{XGB}}}
\newcommand{\Enet}{{\texttt{Elastic-Net}}}
\newcommand{\Ridge}{{\texttt{Ridge}}}
\newcommand{\RNN}{{\texttt{NN}{}}}

\newcommand{\baseline}{baseline}

\newcommand{\no}{no}
\newcommand{\BOTH}{Reg/Classif}
\newcommand{\Clf}{Classif} 
\newcommand{\reg}{Reg} 
\newcommand{\bs}{b_s}

\newcommand{\dither}{dither}
\newcommand{\Dither}{Dither}

\newcommand{\textapprox}{\scalebox{1}{\raisebox{0.7ex}{\texttildelow}}}

\newcommand{\Hd}{{\texttt{Hardmax}}}
\newcommand{\s}{\texttt{Sig}}
\newcommand{\BCE}{{\texttt{BCE}}}
\newcommand{\sigmoid}{{\texttt{Sigmoid}}}
\newcommand{\BMLR}{{\texttt{Ens-MLR}}}
\newcommand{\BMLRa}{{\texttt{Bag-MLR1}}}
\newcommand{\BMLRL}{{\texttt{Bag-MLR$\textapprox$L}}}
\newcommand{\BMLRb}{{\texttt{Bag-MLR2}}}
\newcommand{\MLRa}{{\texttt{MLR$\textapprox$1}}}
\newcommand{\MLRb}{{\texttt{MLR$\textapprox$2}}}
\newcommand{\MLRc}{{\texttt{MLR$\textapprox$3}}}
\newcommand{\MLRd}{{\texttt{MLR$\textapprox$4}}}
\newcommand{\MLRL}{{\texttt{MLR$\textapprox$L}}}
\newcommand{\MLR}{{\texttt{MLR}}}
\newcommand{\BestMLR}{{\texttt{Best-MLR}}}
\newcommand{\TopfiveMLR}{{\texttt{Top5-MLR}}}

\newcommand{\MARS}{{\texttt{MARS}}}
\newcommand{\LM}{{\texttt{LM}}}
\newcommand{\Baseline}{{\texttt{Baseline}}}
\newcommand{\TREE}{{\texttt{TREE}}}
\newcommand{\QDA}{{\texttt{QDA}}}
\newcommand{\GLM}{{\texttt{GLM}}}

\newcommand{\NN}{{\texttt{MLR-NN}}}
\newcommand{\ReLu}{{\texttt{ReLu}}}
\newcommand{\FFNN}{{\texttt{FFNN}}}
\newcommand{\MSE}{{\texttt{MSE}}}
\newcommand{\RMSE}{{\texttt{RMSE}}}
\newcommand{\CE}{\texttt{CE}}
\newcommand{\CEMLR}{\texttt{BCE-MLR}}
\newcommand{\CENN}{\texttt{BCE-MLR-NN}}

\newcommand{\perm}{\pi}

\newcommand{\Input}{{\texttt{input}}}
\newcommand{\out}{{\texttt{out}}}

\newcommand{\lr}{{\texttt{$\ell_r$}}}
\newcommand{\fixB}{{\texttt{FixB}}}
\newcommand{\iter}{{\texttt{Iter}}}
\newcommand{\Iter}{{\texttt{Iter}}}

\newcommand{\If}{{\textbf{if}}}

\newcommand{\While}{\textbf{{while}}}
\newcommand{\Do}{\textbf{{do}}}
\newcommand{\For}{\textbf{{for}}}

\newcommand{\Remove}{\textbf{\texttt{remove}}}

\newcommand{\Set}{\textbf{\texttt{set}}}

\newcommand{\fR}{\textbf{{function-T}}}

\newcommand{\bA}{\boldsymbol{A}}
\newcommand{\bH}{\boldsymbol{H}}
\newcommand{\tbH}{(\I_n-\boldsymbol{H})}

\newcommand{\bx}{\boldsymbol{x}}
\newcommand{\bY}{\boldsymbol{Y}}

\newcommand{\bW}{\boldsymbol{P}}

\newcommand{\bla}{\boldsymbol{\lambda}}
\newcommand{\bth}{\boldsymbol{\theta}}
\newcommand{\bO}{\boldsymbol{0}}

\newcommand{\R}{\mathbb{R}}

\newcommand{\cG}{\ensuremath{\mathcal{G}}}

\newcommand{\cI}{\ensuremath{\mathcal{I}}}

\newcommand{\cN}{\ensuremath{\mathcal{N}}}

\newcommand{\hatla}{\widehat{\boldsymbol{\lambda}}}

\newcommand{\hbth}{\boldsymbol{\widehat{\theta}}}

\newcommand{\1}{\mathbbm{1}}

\newcommand{\LAS}{\texttt{Lasso}}

\newcommand{\I}{\ensuremath{\mathbb{I}}}

\makeatletter
\newcommand*\bigcdot{\mathpalette\bigcdot@{.5}}
\newcommand*\bigcdot@[2]{\mathbin{\vcenter{\hbox{\scalebox{#2}{$\m@th#1\bullet$}}}}}
\makeatother

\newcommand{\el}{\ell}

\newcommand{\eps}{\epsilon}

\newtheorem{mydef}{Definition}

\usepackage{tikz}
\usetikzlibrary{arrows,backgrounds}
\usepgflibrary{shapes.multipart}

\title{Muddling Label Regularization: Deep Learning for Tabular Datasets}
\author{
Karim Lounici\\
  CMAP, Ecole Polytechnique\\
  Route de Saclay\\
  91128  PALAISEAU Cedex\\
  FRANCE\\
\texttt{karim.lounici@polytechnique.edu} \\
Katia Meziani\\
  CEREMADE - Universit\'e Paris Dauphine-PSL\\
  Place du Mar\'echal De Lattre De Tassigny\\
  75775 PARIS CEDEX 16\\
  FRANCE\\ 
\texttt{meziani@ceremade.dauphine.fr} \\
Benjamin Riu\\
  CMAP, Ecole Polytechnique\\
  Route de Saclay\\
  91128  PALAISEAU Cedex\\
  FRANCE\\
\texttt{benjamin.riu@polytechnique.edu}
}

\begin{document}

\maketitle

\vskip 0.3in

\begin{abstract}
Deep Learning (\DL) is considered the state-of-the-art in computer vision, speech recognition and natural language processing. Until recently, it was also widely accepted that \DL{} is irrelevant for learning tasks on tabular data, especially in the small sample regime where ensemble methods are acknowledged as the gold standard.
We present a new end-to-end differentiable method to train a standard \FFNN{}. Our method, \textbf{Muddling labels for Regularization} (\texttt{MLR}), penalizes memorization through the generation of uninformative labels and the application of a differentiable close-form regularization scheme on the last hidden layer during training. \texttt{MLR} outperforms classical \RNN{} and the gold standard (\GBDT, \RF) for regression and classification tasks on several datasets from the UCI database and Kaggle covering a large range of sample sizes and feature to sample ratios. Researchers and practitioners can use \texttt{MLR} on its own as an off-the-shelf \DL{} solution or integrate it into the most advanced ML pipelines. 
\end{abstract}

\section{Introduction} 

Over the last decade, we have witnessed the spectacular performance of Deep Learning (\DL) in the fields of computer vision \cite{Krizhevsky2012_cnn}, audio \cite{Hinton2012} and natural language processing \cite{bert2019}. Until recently, it was also widely believed that \DL{} is irrelevant for tabular data \cite{videolecture}. While the need to handle tabular data arises in many fields ($e.g.$ material science \cite{FENG2019300}, medecine \cite{Rajkomar2018}, online advertising \cite{Zhou2018click,Guo2016}, finance \cite{buehler_deep_2019}), \DL{} for tabular data remains understudied and underused. It may seem strange considering that tabular data appears at first more straightforward to handle than image or textual data.

Most experiments seem to indicate that tree based ensemble methods \cite{Denil2014narrow,Delgado14a,Mentch2020Rand,Wainberg2016}
are the most reliable option on tabular data and often work well even without any parameter tuning \cite{Delgado14a,Wainberg2016}. By contrast, training \DL{} models usually requires extensive trial and error, expertise and time to properly tune the hyperparameters \cite{smith2018disciplined}. This comparison is even worse in the small data regime ($n < 10^3$ or even $n<300$). In many fields including material sciences \cite{FENG2019300}, medicine \cite{JTD4418,Rajkomar2018}, environmental studies \cite{Cassotti2015similarity-based}, small datasets are not rare occurrences since collecting samples and assembling large datasets may be costly, or even impossible by the nature of the task at hand\footnote{$e.g.$ decision making on a limited amount of cases during a pandemic.}.

Deep learning mostly relies on transfer learning schemes to tackle small datasets \cite{Oquab_2014_CVPR}, which are almost never an option for tabular datasets \cite{shavitt2018}. This is due to the lack of transferable domain knowledge between tabular features. Another important line of work focuses on the preprocessing of categorical features which was an historical limitation of \DL{} \cite{qu_product-based_2016, guo_deepfm:_2017, wang_collaborative_2015,cheng_wide_2016,covington_deep_2016,zhang_deep_2019}. In that regard, entity embeddings \cite{Guo2016}, which try to capture relationships between categories, have become standard in \DL{} libraries (tensorflow, PyTorch, \textit{etc.}). Meanwhile, tree based ensemble methods and Gradient Boosting Decision Trees (\GBDT{}) are still considered the best option to handle categorical data \cite{Prokhorenkova2018,Anghel2018Bench,Harasymiv2015_gbdt}.

Developing a \DL{} solution for tabular data is desirable at it can leverage the particular strengths of \DL{}, in particular its ability to perform automatic feature engineering and end-to-end training via gradient descent. The recent \texttt{PyTorch Tabular} project \cite{pytorchtab} and the growing number of articles on tabular data in recent years show the increasing interest of the \DL{} community in this topic \cite{YangMorillo2018,arik2020tabnet,ZhouFeng2017,MillerHHJK17,FengYuZhou2018,ke2019tabnn,Olson2018moderna,Arora2020meuf,katzir2021netdnf,Klambauer2017,shavitt2018,Haldar2019,Popov2020Neural,pmlr-v97-du19a}.

\begin{table}[!hp]
\caption{State-of-the-art: Deep Learning for supervised tasks on tabular data. $\bm{*}$ uses the UCI database which has some well-known flaws \cite{Arora2020meuf,Klambauer2017,Wainberg2016}.}
\label{tab:SotA}
\centering
\resizebox{\columnwidth}{!}{
\begin{tabular}{|l|c|c|c|cc|c|}
\hline
Method        &  End-to-end & Works without & Task & \multicolumn{2}{|c|}{ Benchmark }    & Consistently outperforms  \\
       &   differentiable & HP tuning &  & \# datasets& Range $n$   & \GBDT \\
       \hline
\TabNN{} \cite{ke2019tabnn}             & \no & \no & \BOTH &5&14.8K-7.3M & no\\
\hline
\NODE{} \cite{Popov2020Neural}             & \checkmark & \no & \BOTH &6&500K-11M & no\\
\hline
\TabNet{} \cite{arik2020tabnet}              & self-supervised  & \no  & \BOTH & 4&10K-11M & \checkmark\\
\hline
\DNDT{} \cite{YangMorillo2018}             & \checkmark & \checkmark &\Clf &14& 150-1.1M & no\\
\hline
\NTK{} \cite{Arora2020meuf}            & \checkmark & \no & \Clf &90\textbf{*}& 10-130K & no  \\ 
\hline
\SNN{} \cite{Klambauer2017}          & \checkmark & \no & \BOTH &122\textbf{*}&  10-130K & no\\
\hline
\NetDNF{} \cite{katzir2021netdnf}         &  no & \checkmark & \Clf & 6&9.8K-200K &no\\
\hline

\RLN{} \cite{shavitt2018}             & \checkmark & \checkmark& Reg & 9&2.5K& no \\ \hline

\MLR{} (this work)          & \checkmark & \checkmark & \BOTH & 32 & 72-65K & Reg:\checkmark\\
\hline
\end{tabular}
}
\end{table}

\paragraph{Regularization.} Two classical \DL{} regularization strategies, dropout \cite{srivastava14a} and weight decay \cite{Hanson1988Comparing}, have been compared by \cite{ZhangDW16} on tabular data. They found dropout to be better, however dropout may still fail in some tabular data tasks \cite{Haldar2019}. Moreover \cite{SongS0DX0T19,ZhangDW16,Qu2018,guo_deepfm:_2017} seem to indicate that dropout parameters are data and model dependent.

\paragraph{Interpretability.} 
The "black-box" aspect of \DL{} remains a major obstacle to its wider use as interpretability in AI is an important concern \cite{chen2018learning}. Recently, a line of research focuses on the development of novel network architectures with interpretable features. \DNDT{} \cite{YangMorillo2018} is a specific neural network architecture which can be trained via end-to-end gradient descent. In addition, it can also be rendered as a decision tree for the sake of interpretation. However \DNDT{} is not scalable $w.r.t.$ the number of features and does not outperform Random Forests (\RF) or standard \RNN{} on the UCI database. Attention-Mechanism (\AM) has boosted \DL{} performance on a range of \NLP{} tasks (\cite{Bahdanau2014-Attention,bert2019}). It turns out that \AM{} can also be used for interpretability purpose. Recently \cite{arik2020tabnet} exploited \AM{} to develop \TabNet{}, an interpretable \DL{} method for tabular data, and claimed it outperforms the gold standard on a limited number of data sets of size $\gtrsim 10K$. A limitation of this approach is the complicated data dependent fine-tuning of the hyperparameters.

\paragraph{Hybrid architecture.}  Several recent works propose to combine decision trees with \DL{}. In that regard, \cite{ZhouFeng2017,MillerHHJK17,FengYuZhou2018} proposed to stack layers of \RF{} or \GBDT{}. However these architectures cannot be trained end-to-end, which may result in potentially inferior performance. \TabNN{} \cite{ke2019tabnn} is a hybrid machine learning algorithm using \GBDT{} and Deep Neural Networks (\DNN). \TabNN{} outperforms standard Feed-Forward Neural Networks (\FFNN{}) but the improvement over \GBDT{} seems marginal in their experiments on 6 data sets ranging in size from $15$K up to $7.9$M training samples. More recently, \NODE{} \cite{Popov2020Neural}, a new \DNN{} architecture consisting of differentiable oblivious decision trees, can be trained end-to-end via backpropagation. \NODE{} marginally outperforms ensemble methods (\Catboost \cite{Prokhorenkova2018}, \XGBoost \cite{guestrin2016}) on 4 out of 6 large size tabular data sets
and requires careful hyperparameter optimization.

\paragraph{New loss functions.} Our contribution falls in this line of research. It consists in replacing the usual loss used to train \DNN{} by specific losses with interesting properties. In that regard, Regularization Learning Networks (\RLN) \cite{shavitt2018} is a new family of neural networks trained with a new loss, named counterfactual loss, together with stochastic gradient descent. \RLN{} performs significantly better than standard \RNN{} but could not beat \GBDT{}.

\paragraph{Other approaches.} We can also cite Neural Tangent Kernel (\NTK){}\cite{Arora2020meuf}, \NetDNF{} \cite{katzir2021netdnf} and Self-Normalized Neural Networks (\SNN) \cite{Klambauer2017}.  Table \ref{tab:SotA} summarizes their properties. We provide more details about these methods in the Appendix.

\paragraph{Contributions.} 

We propose a pure deep learning solution to train a standard \FFNN{} for tabular data. Our method, \textbf{Muddling labels for Regularization} (\MLR), penalizes memorization over permuted labels and structured noise through the application of a differentiable close-form regularization scheme on the last hidden layer during training. More specifically:

- Our method outperforms usual methods (Ensemble, \SVM, Boosting, Linear Regression, $etc.$ ) including the gold standards \RF{} and \GBDT{} for the usual statistics (Mean $\Rdeux$, Friedman rank, P90, P95, P98, PMA) on a diverse collection of regression datasets. Our method also comes in a close second for classification tasks.

- The \MLR{} method only requires the most basic standardization, one-hot-encoding and standard imputation of missing data. \MLR{} is fully compatible with all feature engineering schemes ($e.g.$ embeddings, Nystr\"om \cite{Williams2001using} and \RBF{} \cite{RasmussenW06} kernels, tree leaves). All the popular \DL{} schemes can also be leveraged including learning rate schedulers \cite{smith2015cyclical}, optimizers, weight decay, batch-normalization, drop-out, residual layers and leaky activations \cite{smith2018disciplined}. 

- The performances of \NN{} are not tied with any of the well-known class of methods. Thus they should be a great addition to the stack of models aggregated by meta-learners. Researchers and practitioners can use \MLR{} on its own as an off-the-shelf \DL{} solution or integrate it into the most advanced ML pipelines.

- The implementation of our method in torch is available as a stand-alone which follows the scikit-learn API (\textit{i.e.} it can be directly encapsulated into parameter search routines, bagging meta models, \textit{etc.}). For the sake of replicability,  the code to run the benchmarks, the ablation study and the preprocessing applied to each dataset is also provided.

The rest of the paper is organized as follows. 
We describe our approach in Section \ref{sec:MLRmethod}. In Section \ref{sec:exp}, we carry out a detailed ablation study of our method and evaluate its performances on real data.

\section{The \MLR{}-\FFNN{}}
\label{sec:MLRmethod}

\subsection{The (\MLR{}) method for Regression}

Let $\mathcal{D}_{train}=(\bx,\bY)=\{(\bx_i,Y_i)\}_{i=1}^n$ be the $train$-set with $\bx_i\in\R^d$ where $d$ denotes the number of features and $Y_i \in \R$. We consider a simple \FFNN{} with $L$ layers, $J$ nodes on each hidden layer and the \ReLu activation function between each hidden layer. For $n$ observations $\bx$, we set $\bA^0=\bx\in\R^{n\times d}$ and
$\bA^1=\ReLu(\bA^{0}W^{1} +B^{1}),\,\,W^1\in\R^{d\times J}$
\begin{eqnarray}
\label{FFNN}
\bA^{\el+1}&=&\ReLu(\bA^{\el}W^{\el+1} +B^{\el+1}),\quad W^{\el+1}\in\R^{J\times J},\,\,\forall\el\in\llbracket1,L-2 \rrbracket,\nonumber\\
\bA^L&=&\bA^{L-1}W^L,\quad W^L\in\R^{J\times 1},
\end{eqnarray}
where $\forall\el\in\llbracket1,L-1 \rrbracket$, $B^{\el} = \1_n \otimes b^{\el}$, $b^{\el}\in\R^{J}$ are the bias terms.

The 3 essential ingredients of the \MLR{} method are Ridge regularization, structured dithering and random permutations as they promote generalization when we train this \FFNN{}. 

We introduce first the Ridge regularization. For $\bla>0$, we set
\begin{eqnarray}
\label{W}
\bW&=&\bW(\bth,\bla,\bx)=\left[(\bA^{L-1})^\top\bA^{L-1}+\bla \I_J\right]^{-1}(\bA^{L-1})^\top\in\R^{J\times n}\\
\label{H}
\bH&=&\bH(\bth,\bla,\bx)=\bA^{L-1}\bW\in\R^{n\times n}
\end{eqnarray}
where the last hidden layer is $\bA^{L-1}:=\bA^{L-1}(\bth,\bx)$ and $\I_J$ denotes the identity matrix%
. Note that $\bH$ is differentiable w.r.t. $\bth =\{(W^{\el},b^{\el})\}_{\el=0}^L$ and $\bla$. We apply Ridge regularization\footnote{Ridge model :  $f(\bth,\bla,\bx)=\bx\,\beta_{\bla}(\bx,\bY):=\bx\,(\bx^\top\bx+\bla\I)^{-1}\bx^\top\bY$}
to the last hidden layer $\bA^{L-1}$ instead of input $\bx$: 
\begin{eqnarray}
\label{ridge}
f(\bth,\bla,\bx):=\bA^{L-1}W^L=\bA^{L-1}\bW(\bth,\bla,\bx)\bY=\bH(\bth,\bla,\bx)\bY.
\end{eqnarray}

Next we introduce the \textbf{permutations}. For a permutation $\perm$ of $n$ elements, we define the corresponding label permutation operator $\perm$ of $\bY = (Y_1,\cdots,Y_n)$ as
$
\perm(\bY)=(Y_{\perm(1)},\cdots,Y_{\perm(n)}).
$
Fix $T\geq 1$ and draw $T$ label permutation operators \textbf{uniformly at random} in the set of all possible label permutations : $(\perm^t(\bY))_{t=1}^T$. This operation can be seen as a form of data-augmentation on the labels.

\begin{mydef}[\MLR{} regression loss]
\label{MLRlossBigsetting}
Set $\bH=\bH(\bth,\bla,\bx)$. We draw $i.i.d.$ random vectors $\xi$ and $\left(\xi_t\right)_{t=1}^T$ distributed as $\cN(0_n,\I_n)$. Let $\left(\perm^t(\bY)\right)^T_{t=1}$ be $T$ independently drawn permutations of $\bY$. We set $\overline{\bY} = mean(\bY)$ and define the \MLR{} loss as

\begin{align*}
\MLR(\bth,\bla) &= \RMSE\,\left(\bY+\tbH \xi\,;\,\bH\bY\right)\\
&\hspace{1cm}+\frac{1}{T}\sum_{t=1}^T\left|\RMSE\,( \bY\,;\, \overline{\bY}\1_n) -  \RMSE\,\left(\perm^t(\bY)+\tbH \,\xi_t\,;\,\bH\,\perm^t(\bY)\right)\right|.
\end{align*}
\end{mydef}

The \MLR{} loss contains two antagonistic terms and was first introduced in the linear regression setting \cite{MLR}. The first term is the usual \RMSE{} while the second term quantifies the amount of memorization of a model by comparing its \RMSE{} on uninformative labels to the baseline $\RMSE\,( \bY\,;\, \overline{\bY}\1_n)$, $i.e.$ the performance achieved without fitting the data. Using the \RMSE{} instead of the \MSE{} in the comparison slightly improves the generalization performances. We explain below the role of $\tbH \xi$ and $\tbH \xi_t$.

\paragraph{The benefit of close-form regularization.} 

The replacement of the output layer with Ridge regularizes the network in two ways:
$(i)$ the weights on the output layer are a direct function of the last hidden layer $\bA^{L-1}$. This effect is much stronger than adding a constraint or a penalty directly on $W^L$ the weights of the $L$-th layer in (\ref{FFNN}); $(ii)$ the close-form we choose is the Ridge instead of the \OLS, which implicitly subjects the weights to a steerable $L_2$ regularization.

\paragraph{The generalization effect of random permutations.}
Our work is loosely related to \cite{zhang2016understanding} where label permutations are used after the model has been trained as a \textit{qualitative} observational method to exhibit the overfitting capacity of Neural networks. In our approach, we go further as we use random permutations during the training phase to define a \textit{quantitative} measure of the amount of overfitting of a model. More precisely, label permutation is used to produce a control set $(\bx,\perm(\bY))$ that can only be fitted through memorization. \MLR{} focuses on patterns that appear only in $(\bx,\bY)$ and not in uncorrelated pairs $(\bx,\perm(\bY))$. 

\paragraph{Structured Dithering.}

We describe an additional scheme to prevent memorization. We apply a dithering scheme which adapts to the spectral structure of $\bH$, the "regularized projector" based on $\bA^{L-1}$ (the output of the last hidden layer). More specifically, we muddle the target using $\tbH \xi\,$ which introduces noise of higher variance along the weakly informative eigendirections of $\bH$.

\paragraph{Computational point of view.} The permutations are drawn once before the training and are not updated or changed thereafter. Once the \FFNN{} is trained, these permutations have no further use and are thus discarded. In practice we take $T=16$ for all the datasets in our benchmark. Therefore, $T$ \textbf{does not require hyperparameter tuning}. Moreover, note that the choice of the seed used to generate the permutations has no impact on the values of the \MLR{} loss. The additional computational cost of using \MLR{} is marginal. We only need to compute a matrix inverse on the output of the last hidden layer $\bA^{L-1}$. This operation is differentiable and inexpensive as parallelization schemes provide linear complexity on GPU when some memory constraints are met \cite{Murugesan2018Embarrassingly,rajibr2010accelerating,chrzeszczyk2013matrix}.

\subsection{Model: \NN{} and training protocol}

\paragraph{The \NN{} Architecture.} 
We consider the \FFNN{} described in \eqref{FFNN} with $L$ layers and all the hidden layers of constant width $J$. In our experiments, we always take $J$ as large as possible (our machine with 11GVRAM allowed for $J=2^{10}$) and $L\in\cG_L:=\{1,2,3,4\}$. 

The \MLR{} neural net (\NN) is 
\begin{align*}
 \NN(\hbth,\hatla,\bigcdot)=\bA^{L-1}(\hbth,\bigcdot) \,\bW(\hbth,\hatla,\bx) \bY \quad\text{with}\quad (\hbth,\hatla) = \underset{\bth,  \bla}{\arg\min} \;\MLR(\bth,\bla)
\end{align*}
with $\bW(\cdot,\cdot,\bx)$ as defined in \eqref{W}. The initialization of $\bth$ is as in \cite{Gloriotetal}.

\paragraph{Initialization of the Ridge parameter.} 

The initialization of $\bla$ is both crucial and non trivial. Choosing $\bla$ close to $0$ will hinder regularization. Furthermore, a very small value of $\bla$ will cause numerical instability during the matrix inversion. Conversely, choosing $\bla$ too big will prevent any learning. Indeed, the gradient with respect to $\bla$ will vanish in both cases. From our initial calibration, we discovered that there exists no universal value to initialize $\bla$. The appropriate initial value depends on many factors such as data size, network architecture, difficulty of the task, $etc$. However, we found a very efficient heuristic to pick an appropriate initial value. We want to start training from the point where fitting the data will lead to generalization as much as possible instead of memorization. In this region, the variation of \MLR{} with respect to $\bla$ is maximum. In practice, we pick $\bla_{init}$ by running a grid-search on the finite difference approximation for the derivative of \MLR{} in (\ref{lambdainiti}) on the grid 
$\cG_{\bla}=\left\lbrace \bla^{(k)} = 10^{-1}\times10^{5 \times k / 11}\;: \;   k = 0, \cdots, 11 \right\rbrace$:
\begin{eqnarray}
\label{lambdainiti}
\bla_{init}=\sqrt{\bla^{(\hat {k}})\,\bla^{(\hat {k}+1)}}\;\text{where}\;\hat {k}=\arg\max\left\{ \left(\MLR(\bth,\bla^{(k+1)}) - \MLR(\bth,\bla^{{(k)}})\right),\, \bla^{(k)}\in\cG_{\bla}\right\}.
\end{eqnarray}
From a computational point of view, the overcost of this step is marginal because we only compute $\bA^{L-1}$ once, and we do not compute the derivation graph of the $11$ matrix inversions or of the unique forward pass. The Ridge parameter $\bla$ is \textbf{not an hyperparameter} of our method; it is trained alongside the weights of the Neural Net architecture. 

\paragraph{\Dither{} \cite{dither}.} 
We do not apply the \MLR{} loss on $\bY$ and the permuted labels $\left(\perm^t(\bY)\right)_{t=1}^T$ but rather on noisy versions of them. We draw $T+1$ $i.i.d.$ $\cN(\bO_n,\widetilde\sigma^2\I_n)$ noise vectors that are added to $\bY$ and $\perm^t(\bY)$, $1\leq t \leq T$. Here again, $\widetilde\sigma$ is \textbf{not an hyperparameter} as we use the same value $\widetilde\sigma=0.03$ for all the data sets in our benchmark.

\paragraph{Training protocol.}

Using wider architecture ($J=2^{10}$) and bigger batch size ($\bs = \min(J,n)$) is always better. To train our \FFNN{}, we use Adam \cite{kingma2014adam} with default parameters except for the learning rate $\lr$ (which depends on the number of layers $L$. See Table~\ref{tab:architectures1}) and we select a $validation$-set of size $n_{val}=20\%\, n$.

{\it Choice of $\max_{\iter}$ and early stopping.} We fix the budget (\fixB = 5 min) and denote by 
$n_{\iter}$ the possible number of iterations during the alloted time \fixB. We fix the maximum number of iterations $\max_{\iter}$ (depending on the value of $L$).
Then, $\iter = \min(\max_{\iter},n_{\iter})$ is the number of iterations that will actually be performed.
We read the $\Rdeux$-score for each iteration on the $validation$-set and take the iteration with the best $\Rdeux$-score: 
$\iter^*:=\arg\max\{\texttt{R}^2_k, \ k =1,\cdots,\iter\}$
. Finally, $(\hbth,\hatla)$ will take its value at iteration $\iter^*$.

\begin{table}[!hp]
	\caption{Benchmarked architectures. }
	\label{tab:architectures1}
\footnotesize
 	\centering
 \begin{tabular}{|c||c|c |c|c|c |c|c|c|c|}
 \hline
Architecture & $L$ & $\lr$ & $\max_{\iter}$ & \fixB & $J$ & $\bs$ & $T$ & \multicolumn{2}{|c|} {$\widetilde\sigma$ }   \\ 
\hline
 $\MLRa $& $1$ & $10^{-2}$ & $200$ & $10'$ & $2^{10}$ & $min(n, J)$ & $16$ & Reg.: $0.03$ & Classif.: $0$ \\
  \cline{9-10} 
 $\MLRb $& $2$ & $10^{-3}$ & $200$ & $id.$ & $id.$ & $id.$& $id.$ &\multicolumn{2}{|c|} {$id.$}\\
 $\MLRc $& $3$ & $10^{-3.5}$ & $400$ &  $id.$ & $id.$ & $id.$& $id.$ & \multicolumn{2}{|c|} {$id.$}\\
 $\MLRd $& $4$ & $10^{-4}$ & $400$ &  $id.$ & $id.$ & $id.$& $id.$ & \multicolumn{2}{|c|} {$id.$} \\
 \hline
 \end{tabular}

 \end{table}

The generic values ($\tilde{\sigma} = 0.03$ and $T=16$) for the \dither{} and the number of permutations hyperparameters yield consistently good results overall. The \dither{} parameter admits an optimal value which seems to correspond to the standard deviation of the target noise. As soon as $T=16$, the choice of permutations has little impact on the value and the \MLR{} loss. In addition, when $T=16$, GPU parallelization is still preserved.
Recall that the Ridge parameter $\bla$ is trained alongside the weights of the \FFNN{} architecture and the initial value $\bla_{init}$ is fixed by the heuristic choice (\ref{lambdainiti}). Our investigations reveals that this choice is close to the optimal oracle choice on the test set. We can also see that the runtime overhead cost of replacing a matrix multiplication with a matrix inversion depends only linearly on the width $J$ and the batch size $\bs$, which are fixed in our method. As a pure DL method, \MLR{} method is scalable. Its complexity is the same as training a standard \RNN{} \cite{Murugesan2018Embarrassingly}. We refer to the Appendix for a detailed description of the training protocol.

\paragraph{Our final models.}

We propose several models with varying depth based on \FFNN{} trained with the \MLR{} loss. We also create ensemble models combining architectures of different depth.
Our models are:\\
$\bullet$  \MLRL: a simple \FFNN{} of depth $L$ ($1\leq L\leq 4$).\\
$\bullet$ \BMLRL: a bagging of 10 \FFNN{} of depth $L$ ($L=1$ or $L=2$).\\
$\bullet$ \BMLR{}: an ensemble of 20 \FFNN{} (the aggregation of \BMLRa{} and \BMLRb{} of depth $1$ and $2$ respectively).\\
$\bullet$ \BestMLR{}: the best prediction among 20 \NN{} in terms of the validation score.\\
$\bullet$ \TopfiveMLR{}: the aggregation of the top 5 among 20 \NN{} in terms of the validation score.

For the methods based on bagging \cite{breiman1996}, the final prediction is the mean of each \NN{} prediction.

\subsection{Classification tasks with the \CEMLR{} loss}

The adaptation of the \MLR{} method to classification tasks is relatively simple. The \FFNN{} architecture and the training protocol are essentially unchanged. The usual loss for binary classification task is the \BCE{} loss that combines a Sigmoid and the Cross Entropy (\CE) loss. Set $\s(\cdot)=\sigmoid(\cdot)$, then
$$\BCE(\bY, f(\bth,\bx))=-\frac1n\left[\bY^\top\log(\s(f(\bth,\bx))
+(\1_n-\bY)^\top\log(\1_n-\s(f(\bth,\bx))\right].$$

\begin{mydef}[\CEMLR{} loss]
\label{CElossBigsetting}
Let $\xi$ and $\left(\xi_t\right)_{t=1}^T$ be $i.i.d.$ $\cN(0_n,\I)$ vectors. Set $\bY^* = 2 \bY - 1$. We define the \CEMLR{} loss as
\begin{align*}
\CEMLR(\bth,\bla) &= \BCE\,\left(\bY;\,\bY^*+\tbH \xi+\bH\bY^*\right)
\\
&\hspace{0.25cm}+\frac{1}{T}\sum_{t=1}^T\left|\BCE\,( \bY\,;\, \overline{\bY}\1_n) -\BCE\,\left(\perm^t(\bY^*) \,;\,  \perm^t(\bY^*)+\tbH \xi_t+\bH\,\perm^t(\bY^*)\right)\right|.
\end{align*}
\end{mydef}

The quantity $\BCE\,( \bY\,;\, \overline{\bY}\1_n)$ is our \baseline. The structured dithering is applied to the prediction rather than the target $\bY$ because the \BCE{} is only defined for binary target $\bY \in \{0;1\}^n$.

The \CEMLR{} neural net (\CENN) is 
\begin{align*}
    \label{CENN}
 &\CENN(\hbth,\hatla,\bigcdot)=\Hd(\bA^{L-1}(\hbth,\bigcdot) \,\bW(\hbth,\hatla,\bx)) \bY \in\{0,1\}^{obs.}\noindent\\
&\text{with}\,\, (\hbth,\hatla) = \underset{\bth,  \bla}{\arg\min} \;\CEMLR(\bth,\bla),
\end{align*}
with $\bW(\cdot,\cdot,\bx)$ defined as in \eqref{W}. 
We refer to the Appendix for a detailed discussion on this specific adaptation.

\section{Experiments}\label{sec:exp}

We provide both the code to download raw files and apply each steps, and the resulting data matrices. All results are fully reproducible as both random seeds and random states were manually set and saved at every step of our experiments. 

See the supplementary material for the the github repository, the detailed description of our experimental setting and the exhaustive list of compared methods with their performances.

\subsection{Setting.}

\paragraph{Benchmark description.}

To produce this benchmark we aggregated 32 tabular datasets (16 in regression  and 16 in classification), from the UCI repository and Kaggle. 
For computational reasons, we have chosen to restrict the number of datasets but we performed more $train$/$test$ splitting in order to reduce the variance of our results. 
We curated the UCI repository and Kaggle through a set of rules detailed in the appendix ($e.g.$ discard empty or duplicate datasets, times series, missing target, non $i.i.d.$ samples, text format, $etc$.).

\paragraph{Preprocessing.} To avoid biasing the benchmark towards specific methods and to get a result as general as possible, we only applied as little preprocessing as we could, without using any feature augmentation scheme. The goal is not to get the best possible performance on a given dataset but to compare the methods on equal ground. We first removed features with constant values such as sample index. Categorical features with more than 12 modalities were discarded as learning embeddings is out of the scope of this benchmark. We also removed samples with missing target. Next, all missing values are imputed with the mean and the mode for numerical and categorical features respectively. We applied one-hot-encoding for categorical values and standardization for numerical features and target. 

We repeated our experiments 10 times using a different $80:20$ $train$/$test$ split of the data and no stratification scheme.

\paragraph{Compared methods.}
We ran the benchmark with \textbf{all the methods available in the scikit-learn library} for classification and regression (including \RF{} and \XGB) as well as the \GBDT{} methods. In the rest of the paper, we only display the main classes of methods in Table \ref{tab:methods}.

\begin{table}[H]
\caption{Main classes of methods.}
\label{tab:methods}
	\centering
	\footnotesize
\begin{tabular}{|c|c|}
\hline
Class      & \\
of Methods & Methods\\
\hline
\MLR{} (this paper)   & \MLRL, \BMLRL, \BMLR, \BestMLR, \TopfiveMLR  \\
\hline
\GBDT  & \XGB{} \cite{Breiman1997_arcing,Friedman2001,Friedman2002},  \Catboost{} \cite{Prokhorenkova2018}, \XGBoost{} \cite{guestrin2016}, \lightgbm{} \cite{Ke2017} \\
\hline
\RF  & \RF{} and \XRF{} \cite{breiman2001,barandiaran1998random}  \\
\hline               
\SVM &  \texttt{Lin-}\SVM{}, \SVM{}, $\nu$\texttt{-}\SVM{} \cite{Chihchung2011}\\
\hline
\RNN   & \texttt{Fast.ai} \cite{Howard2020}, \MLP \cite{Hinton89connectionistlearning}\\
\hline
\GLM   & \texttt{OLS}, \Enet{} \cite{Zou05}, \Ridge{} \cite{Hoerl1970}, \LAS{} \cite{tibshirani1996}, \texttt{Logistic regression} \cite{cox1958}\\
\hline
\MARS & \MARS{} \cite{Friedman1991}\\
\hline
\TREE& \texttt{CART}, \texttt{XCART} \cite{Breiman1984,gey2005model,Klusowski2020sparse}\\
\hline
\Baseline & Reg: \texttt{Intercept}$\,|\,$ Classif: \texttt{Class probabilities}\\
\hline
\end{tabular}
\end{table}

\subsection{Ablation Analysis.}

\begin{table}[H]
\caption{Ablation Study in Regression. 
}
\label{tab:ablation}
	\centering
	\footnotesize
\begin{tabular}{|l||c|c|}
\hline
Step  &  Mean $\Rdeux$  & Bagging $\Rdeux$ \\
\hline
\hline
\FFNN  & $ -0.081  \pm  0.173 $ & $ -0.046  \pm  0.169 $ \\
\hline
+ Ridge  & $ 0.321  \pm  0.081 $ & $ 0.394  \pm  0.052 $ \\
\hline
+ Ridge + Struct. Dithering  & $ 0.323  \pm  0.075 $ & $ 0.400  \pm  0.048 $ \\
\hline
+ Ridge + Permut. & $ 0.364  \pm  0.050 $ & $ 0.432  \pm  0.035 $ \\
\hline
\MLR{}   & $ 0.371  \pm  0.024 $ & $ 0.433  \pm  0.000 $ \\
\hline
\end{tabular}
\end{table}

We ran our ablation study (Table \ref{tab:ablation}) in the regression setting on 3 datasets with different sample sizes and feature to sample ratios. We repeated each experiment over 100 random $train$/$test$ splits. All the results presented here correspond to the architecture and hyperparameters of \MLRb{} and \BMLRb{}.

A standard \FFNN{} of $2$ layers with a wide architecture ($J=1024$) cannot be trained efficiently on such small datasets as the \FFNN{} instantly memorizes the entire dataset. This cannot be alleviated through bagging at all. Note also its lower overall performance on the complete benchmark (Table \ref{tab:perfR2_rank}). Applying Ridge on the last hidden layer allows an extremely overparametrized \FFNN{} to learn but its generalization performance is still far behind the gold standard \RF{}. However, when using bagging with ten such models, we reach very competitive results, underlying the potential of the \MLR{} approach.

The random permutations component gives a larger improvement than Structured Dithering. However, when using both ingredients together, a single \NN{} can reach or even outperform the gold-standard methods on most datasets. Furthermore, the improvement yielded by using bagging ($0.062$) is still of the same order of magnitude as the one we got when we applied permutations on top of Ridge to the \FFNN{} ($0.043$). This means these two ingredients (permutations and struct. dithering) are not just simple variance reduction techniques but actually generate more sophisticated models.

\subsection{Overall Performance comparisons.}

\begin{table}[H]
\caption{\textbf{Performances of the best method in each class of methods} for the regression task on our benchmark. \Pa, \Pb, \Pc: the number of datasets a model achieves 90\%, 95\%, 98\% or more of the maximum test $\Rdeux$-score respectively, divided by the total number of datasets. \PMA: average percentage of the maximum test $\Rdeux$-score.}
\label{tab:perfR2_rank}
	\centering
	\resizebox{\columnwidth}{!}{
\begin{tabular}{|l|c|c|c|c|c|c|}
\hline
Class      &   &           &               &    &           &                          \\
 of Methods        & \FR  & Mean $\Rdeux$-score          & \Pa              & \Pb    & \Pc          & \PMA                          \\
  \hline
\hline
\hline
\MLR & $2.525 \pm 1.355$ & $0.744 \pm 0.022$ & $0.963$ & $0.856$ & $0.719$ & $0.946 \pm 0.089$\\
\GBDT & $2.719 \pm 1.850$ & $0.726 \pm 0.093$ & $0.863$ & $0.756$ & $0.650$ & $0.898 \pm 0.237$\\
\RF & $3.538 \pm 1.896$ & $0.724 \pm 0.070$ & $0.825$ & $0.681$ & $0.481$ & $0.914 \pm 0.159$\\
\SVM & $4.281 \pm 1.534$ & $0.711 \pm 0.068$ & $0.831$ & $0.594$ & $0.362$ & $0.882 \pm 0.172$\\
\RNN & $4.331 \pm 2.206$ & Aberating value & $0.725$ & $0.606$ & $0.475$ & Aberating value \\
\MARS & $5.644 \pm 1.623$ & $0.677 \pm 0.066$ & $0.537$ & $0.350$ & $0.163$ & $0.861 \pm 0.167$\\
\LM & $5.938 \pm 1.804$ & $0.658 \pm 0.094$ & $0.531$ & $0.294$ & $0.156$ & $0.837 \pm 0.179$\\
\TREE & $7.125 \pm 1.613$ & $0.512 \pm 0.237$ & $0.338$ & $0.188$ & $0.119$ & $0.578 \pm 0.570$\\
\Baseline & $8.900 \pm 0.375$ & $-0.023 \pm 0.211$ & $0.000$ & $0.000$ & $0.000$ & $-0.031 \pm 0.075$\\
\hline
\end{tabular}
}
\end{table}

\textbf{The \MLR{} method clearly outperforms all the compared methods for the regression task.} \BMLR{} with a \Pc{} of $0.719$ on the whole benchmark and Friedman Rank of $2.525$ is above \GBDT{}, with a \Pc{} of $0.65$ and Friedman Rank $2.719$ in Table \ref{tab:perfR2_rank}. 
As revealed by its \PMA{} statistics at $0.946$
, \BMLR{} is far ahead of the other methods. 
This means that \MLR{} produces reliable results at a rate that is even above methods like \RF{} which are often deemed the safest pick. 
Standard \RNN{} with equivalent architecture and \MSE{} loss performs poorly with a Friedman rank of $4.331$. Noticeably, \BMLR{} was most often the best method among all the \MLR{} methods.

\begin{table}[H]
\caption{\textbf{Performances of the best method in each class of methods} for the classification task with the accuracy score.}
\label{tab:ACCperf}
	\centering
	\resizebox{\columnwidth}{!}{
\begin{tabular}{|l|c|c|c|c|c|c|}
\hline
Class      &   &           &               &    &           &                          \\
 of Methods        & \FR  & Mean \Acc        & \Pa              & \Pb    & \Pc          & \PMA                          \\ \hline
\hline
\hline
\GBDT & $1.769 \pm 0.998$ & $0.889 \pm 0.038$ & $0.963$ & $0.881$ & $0.819$ & $0.971 \pm 0.054$\\
\MLR & $2.913 \pm 1.403$ & $0.882 \pm 0.031$ & $0.963$ & $0.869$ & $0.800$ & $0.956 \pm 0.070$\\
\RF & $3.056 \pm 1.415$ & $0.882 \pm 0.038$ & $0.912$ & $0.819$ & $0.656$ & $0.958 \pm 0.063$\\
\GLM & $3.756 \pm 1.561$ & $0.862 \pm 0.060$ & $0.806$ & $0.631$ & $0.463$ & $0.940 \pm 0.062$\\
\TREE & $4.763 \pm 1.195$ & $0.836 \pm 0.062$ & $0.731$ & $0.381$ & $0.237$ & $0.908 \pm 0.084$\\
\QDA & $5.675 \pm 1.688$ & $0.723 \pm 0.160$ & $0.338$ & $0.194$ & $0.169$ & $0.796 \pm 0.159$\\
\Baseline & $6.856 \pm 1.574$ & $0.593 \pm 0.168$ & $0.069$ & $0.025$ & $0.025$ & $0.661 \pm 0.133$\\
\RNN & $7.213 \pm 0.980$ & $0.565 \pm 0.152$ & $0.025$ & $0.013$ & $0.013$ & $0.625 \pm 0.136$\\
\hline
\end{tabular}
}
\end{table}

\begin{table}[H]
\caption{\textbf{Performances of the best in each class of methods} for the classification task with \AUC{} score.}
\label{tab:AUCperf}
	\centering
	\resizebox{\columnwidth}{!}{
\begin{tabular}{|l|c|c|c|c|c|c|}
\hline
Class      &   &           &               &    &           &                          \\
of Methods        & \FR  & Mean \AUC        & \Pa              & \Pb    & \Pc          & \PMA                          \\ \hline
\hline
\hline
\GBDT & $1.738 \pm 1.190$ & $0.918 \pm 0.048$ & $0.938$ & $0.912$ & $0.875$ & $0.963 \pm 0.108$\\
\MLR & $2.900 \pm 1.304$ & $0.908 \pm 0.012$ & $0.912$ & $0.844$ & $0.694$ & $0.952 \pm 0.106$\\
\RF & $2.938 \pm 1.390$ & $0.912 \pm 0.047$ & $0.931$ & $0.887$ & $0.706$ & $0.956 \pm 0.095$\\
\LM & $3.881 \pm 1.572$ & $0.889 \pm 0.060$ & $0.775$ & $0.662$ & $0.475$ & $0.935 \pm 0.094$\\
\RNN & $4.856 \pm 1.545$ & $0.843 \pm 0.154$ & $0.706$ & $0.506$ & $0.412$ & $0.896 \pm 0.155$\\
\TREE & $5.975 \pm 1.160$ & $0.813 \pm 0.091$ & $0.394$ & $0.212$ & $0.212$ & $0.852 \pm 0.119$\\
\QDA & $6.031 \pm 1.371$ & $0.772 \pm 0.149$ & $0.394$ & $0.256$ & $0.150$ & $0.818 \pm 0.152$\\
\Baseline & $7.681 \pm 1.084$ & $0.499 \pm 0.151$ & $0.006$ & $0.000$ & $0.000$ & $0.537 \pm 0.072$\\
\hline
\end{tabular}
}
\end{table}

For binary classification task with the usual accuracy score, \MLR{} is a close second behind \GBDT{} both in terms of Accuracy and \AUC{} scores.

\section{Conclusion}

All these findings reveal \MLR{} as a remarkably reliable method for tabular datasets, one which consistently produces either state-of-the-art or very competitive results, for a large range of sample sizes, feature to sample ratios, types of features and difficulty across very diverse areas of applications. Furthermore, \MLR{} can achieve these steady performances without any intensive tuning. Nonetheless, higher performances can be achieved with the \MLR{} approach by data-dependent tuning of the hyperparameters in Table \ref{tab:architectures1} and/or leveraging usual \DL{} schemes.

By replacing the standard losses by the \MLR{} loss to train a simple \FFNN{}, we were able to break down the tabular data deadlock and outperform the gold standard. However, nothing in our method is constrained to this setting. The \MLR{} approach is perfectly applicable on \CNN{} for classification tasks in the low sample regime with robustness issues.

\section*{Supplementary Material}

\subsection*{Replicability}

Our Python code is released as an open source package for replication:
\href{https://github.com/anonymousNeurIPS2021submission5254/SupplementaryMaterial}{github/anonymousNeurIPS2021submission5254/}.

\subsection*{Configuration machine}

We ran our experiments using several setups and GPU's: 
\begin{itemize}
    \item Google Cloud Plateform: NVIDIA Tesla P100,
    \item Google Colab : NVIDIA Tesla TESLA K80 and NVIDIA Tesla TESLA T4,
    \item Personal Computer : NVIDIA RTX 2080 Ti and NVIDIA RTX 2080 MaxQ.
\end{itemize}

\section{State of the Art}
We complete here the review of the existing literature on deep learning on tabular data.


An interesting line of research proposes to transpose the "leverage weak learners" idea underlying ensemble methods into neural networks. \cite{Olson2018moderna} proposes an interpretation of fitted \FFNN{} as ensembles of relatively weakly correlated, low-bias sub-networks. Thus this paper provides some insight on the generalization ability of overparametrized \FFNN{} on small datasets. Their experiments concerns binary classification on the UCI dataset but they did not attempt to outperform ensemble methods as it was not the goal of this work. 

The paper \cite{Arora2020meuf} carried out a study of Neural Tangent Kernel (\NTK) induced by infinitely wide neural networks on small classification tasks. \NTK{} slightly outperforms \RF{} implemented as in \cite{Delgado14a} on small UCI data sets ($n\leq 5000$). \NTK{} performs well on small size ($n \leq 640$) subsets of the CIFAR-10 benchmark but is inferior to ResNet-34 for larger size. However their architecture does not cover the regression task. Moreover, the super-quadratic running time of \NTK{} limits its use in large scale learning tasks.

\NetDNF{} \cite{katzir2021netdnf} is an end-to-end \DL{} model to handle tabular data. Its architecture is designed to emulate Boolean formulas in decision making. However, \XGBoost{}  outperforms \NetDNF{} in their experiments.

\cite{Klambauer2017} proposes Self-Normalized Neural Networks (\SNN) based on the \SELU{} activation function to train very deep feed-forward neural networks more efficiently. \SNN{} architecture is motivated as it makes SGD more stable. However \SNN{} requires careful tuning of hyperparameters and does not outperform \SVM{} or \RF{} on the UCI database.

\section{The \MLR-\FFNN{}}\label{secMLRloss}

\subsection{The \MLR{} loss}

Recall 
$$\bH=\bH(\bth,\bla,\bx)=\bA^{L-1}\left[(\bA^{L-1})^\top\bA^{L-1}+\bla \I\right]^{-1}(\bA^{L-1})^\top\in\R^{n\times n},$$
Where $\bA^{L-1}$ denotes the last hidden layer.
\begin{mydef}[\MLR{} regression loss]
Set $\bH=\bH(\bth,\bla,\bx)$. We draw $i.i.d.$ random vectors $\xi$ and $\left(\xi_t\right)_{t=1}^T$ distributed as $\cN(0_n,\I_n)$. Let $\left(\perm^t(\bY)\right)^T_{t=1}$ be $T$ independently drawn permutations of $\bY$. We set $\overline{\bY} = mean(\bY)$ and define the \MLR{} loss as

\begin{align*}
\MLR(\bth,\bla) &= \RMSE\,\left(\bY+\tbH \xi\,;\,\bH\bY\right)\\
&\hspace{1cm}+\frac{1}{T}\sum_{t=1}^T\left|\RMSE\,( \bY\,;\, \overline{\bY}\1_n) -  \RMSE\,\left(\perm^t(\bY)+\tbH \,\xi_t\,;\,\bH\,\perm^t(\bY)\right)\right|.
\end{align*}
\end{mydef}

\paragraph{The benefit of close-form regularization.} 

The replacement of the output layer with Ridge regularizes the network in two ways:
$(i)$ the weights on the output layer are a direct function of the last hidden layer $\bA^{L-1}$. This effect is much stronger than adding a constraint or a penalty directly on $W^L$ the weights of the $L$-th layer of the \FFNN; $(ii)$ the close-form we choose is the ridge instead of the OLS, which implicitly subjects the weights to a steerable $L_2$ regularization.

\paragraph{The generalization effect of random permutations.} Our work is loosely related to \cite{zhang2016understanding} where label permutations are used after the model has been trained as a \textit{qualitative} observational method to exhibit the overfitting capacity of Neural networks. In our approach, we go further as we use random permutations during the training phase to define a \textit{quantitative} measure of the amount of overfitting of a model. This measure is actively used to penalize overfitting during the training phase. This is the underlying mecanism behind the \MLR{} loss.
First, when we take a permuted label vector we obtain a new label vector with two properties. First both $\bY$ and $\perm(\bY)$ admit the same marginal distributions. This new vector can be seen as a "realistic" data-augmented new sample for the training set. Second the expected number of fixed points ($\perm(i) = i$) in a permutation drawn uniformly at random is equal to $1$ (See Chapter 5 in \cite{permutebook}); $i.e.$ the proportion of fixed points in a random permutation of $n$ elements is insignificant. Thus the label permutation breaks the dependence relationship between $Y_{\perm(i)}$ and $\bx_i$. Therefore, $\bx_i$ provides no information on the possible value of $Y_{\perm(i)}$ and predicting $Y_{\perm(i)}$ using $\bx_i$ can only result in overfitting. In other words, label permutation is used to produce a control set $(\bx,\perm(\bY))$ that can only be fitted through memorization. \MLR{} focuses on patterns that appear only in $(\bx,\bY)$ and not in uncorrelated pairs $(\bx,\perm(\bY))$.

\paragraph{Structured Dithering.} 

We describe an additional scheme to prevent memorization. We apply a dithering scheme which adapts to the spectral structure of $\bH$, the "regularized projector" based on $\bA^{L-1}$ (the output of the last hidden layer). More specifically, we muddle the target using $\tbH \xi\,$ which introduces noise of higher variance along the weakly informative eigendirections of $\bH$.

\subsection{Cross-Entropy loss}

In the classification task, the \FFNN{} architecture is essentially unchanged. The usual loss for binary classification task is the \BCE{} loss that combines a \sigmoid{} and the Cross Entropy (\CE) loss.
(namely \texttt{torch.nn.BCEWithLogitsLoss} in PyTorch and referred to as \BCE{} in this paper). 
Set $\s(\cdot)=\sigmoid(\cdot)$, then
$$\BCE(\bY, f(\bth,\bx))=-\frac1n\left[\bY^\top\log(\s(f(\bth,\bx))
+(\1_n-\bY)^\top\log(\1_n-\s(f(\bth,\bx))\right].$$

\bigskip

\begin{mydef}[\CEMLR{} loss]
Let $\xi$ and $\left(\xi_t\right)_{t=1}^T$ be $i.i.d.$ $\cN(0_n,\I)$ vectors. Set $\bY^* = 2 \bY - 1$. We define the \CEMLR{} loss as
\begin{align*}
\CEMLR(\bth,\bla) &= \BCE\,\left(\bY;\,\bY^*+\tbH \xi+\bH\bY^*\right)
\\
&\hspace{0.25cm}+\frac{1}{T}\sum_{t=1}^T\left|\BCE\,( \bY\,;\, \overline{\bY}\1_n) -\BCE\,\left(\perm^t(\bY^*) \,;\,  \perm^t(\bY^*)+\tbH \xi_t+\bH\,\perm^t(\bY^*)\right)\right|.
\end{align*}
\end{mydef}

\bigskip

The quantity $\BCE\,( \bY\,;\, \overline{\bY}\1_n)$ is our \baseline. Note that $\bY^*$ with values in $\{-1,1\}$ is the symmetrized version of $\bY$. Next, the Structured dithering is applied to the prediction rather than the target $\bY$ because the \BCE{} is only defined for binary target $\bY \in \{0;1\}^n$. 

\begin{mydef}[\CENN]
Our \CEMLR{} neural net (\CENN) is 
\begin{align*}
 &\CENN(\hbth,\hatla,\bigcdot)=\Hd(\bA^{L-1}(\hbth,\bigcdot) \,\bW(\hbth,\hatla,\bx)) \bY \in\{0,1\}^{obs.}\noindent\\
&\text{with}\,\, (\hbth,\hatla) = \underset{\bth,  \bla}{\arg\min} \;\CEMLR(\bth,\bla),
\end{align*}
and $\bW(\cdot,\cdot,\bx)$ $s.t.$
, $\bW=\bW(\bth,\bla,\bx)=\left[(\bA^{L-1})^\top\bA^{L-1}+\bla \I\right]^{-1}(\bA^{L-1})^\top\in\R^{J\times n}$. 
\end{mydef}

\bigskip

\section{Training a \FFNN{} with \MLR}

\paragraph{The \NN{} Architecture.} 
We consider \FFNN{} with $L$ layers, $L\in\cG_L:=\{1,2,3,4\}$, and with all the hidden layers of constant width $J$. In our experiments, we always take $J$ as large as possible (our machine with 11GVRAM allowed for $J=2^{10}$).

\begin{table}[H]
	\caption{Benchmarked architectures. }
\footnotesize
 	\centering
 \begin{tabular}{|c||c|c |c|c|c |c|c|c|c|}
 \hline
Architecture & $L$ & $\lr$ & $\max_{\iter}$ & \fixB & $J$ & $\bs$ & $T$ & \multicolumn{2}{|c|} {$\widetilde\sigma$ }   \\ 
\hline
 $\MLRa $& $1$ & $10^{-2}$ & $200$ & $10'$ & $2^{10}$ & $min(n, J)$ & $16$ & Reg.: $0.03$ & Classif.: $0$ \\
  \cline{9-10} 
 $\MLRb $& $2$ & $10^{-3}$ & $200$ & $id.$ & $id.$ & $id.$& $id.$ &\multicolumn{2}{|c|} {$id.$}\\
 $\MLRc $& $3$ & $10^{-3.5}$ & $400$ &  $id.$ & $id.$ & $id.$& $id.$ & \multicolumn{2}{|c|} {$id.$}\\
 $\MLRd $& $4$ & $10^{-4}$ & $400$ &  $id.$ & $id.$ & $id.$& $id.$ & \multicolumn{2}{|c|} {$id.$} \\
 \hline
 \end{tabular}
 \end{table}

\paragraph{\Dither{} \cite{dither}.} 
This step is distinct from the Structured dithering that we introduced in the \MLR{} method.
In the regression setting, we do not apply the \MLR{} loss on $\bY$ but rather on a noisy version of $\bY$ as is usually done in practice. Let $\epsilon,\,(\epsilon^t)_t\,\overset{i.i.d}{\sim}\cN(\bO,\tilde\sigma^2\I)$. We set $\bY_{\eps}=\bY+\eps$ and
$\perm_{\epsilon}^t(\bY)=\perm^t(\bY)+\epsilon^{t}$.
In our experiments, we use the \MLR{} loss on $\left(\bY_{\epsilon},\left(\perm_{\epsilon}^t(\bY)\right)_{t=1}^T\right)$ instead of $\left(\bY,\left(\perm^t(\bY)\right)_{t=1}^T\right)$.
\begin{align*}
\MLR(\bth,\bla) &= \RMSE\,\left(\bY_{\eps}+\tbH \xi\,;\,\bH\bY_{\eps}\right)\\
&\hspace{1cm}+\frac{1}{T}\sum_{t=1}^T\left|\RMSE\,( \bY\,;\, \overline{\bY}\1_n) -  \RMSE\,\left(\perm_{\eps}^t(\bY)+\tbH \,\xi_t\,;\,\bH\,\perm_{\eps}^t(\bY)\right)\right|.
\end{align*}

Here again, $\tilde\sigma$ is \textbf{not an hyperparameter} as we use the same value $\tilde\sigma=0.03$ for all the datasets in our benchmark.  Moreover, in our approach the batch size $\bs$ is not a hyperparameter as we fix it as in table above.

Note that we do not apply this dither step in the classification setting.

\paragraph{Initialization of $\bth$.}
 The initialization of $\bth$ is as in \cite{Gloriotetal}.

\medskip

\begin{mdframed}
\underline{Recall $|\Input|=d$ and $|\out|$=1}
\medskip

$\forall\el\in\llbracket1,L-1 \rrbracket$, $b^{\el}=\bO$. The entries of $W^\el$ are generated independently from the uniform distribution on the interval $\cI_\ell$ :  
\begin{itemize}
   \item[$\bullet$] $\cI_1=\left(-\sqrt{\frac{6}{(d+J)}},\sqrt{\frac{6}{(d+J)}}\right)$ and $\,\cI_L=\left(-\sqrt{\frac{6}{(d+1)}},\sqrt{\frac{6}{(d+1)}}\right)$
   \item[$\bullet$] $\cI_\el=\left(-\sqrt{\frac{6}{(J+J)}},\sqrt{\frac{6}{(J+J)}}\right)$, $\forall\el\in\llbracket2,L-1 \rrbracket $
\end{itemize}
\end{mdframed}

\bigskip

\paragraph{Efficient heuristic to initialize the Ridge parameter.} 
In our experiments, we pick $\bla_{init}$ by running a grid-search on the finite difference approximation for the derivative of \MLR{} on the grid 
$\cG_{\bla}=\left\lbrace \bla^{(k)} = 10^{-1}\times10^{5 \times k / 11}\;: \;   k = 0, \cdots, 11 \right\rbrace$:
\begin{eqnarray*}
\bla_{init}=\sqrt{\bla^{(\hat {k}})\,\bla^{(\hat {k}+1)}}\;\text{where}\;\hat {k}=\arg\max\left\{ \left(\MLR(\bth,\bla^{(k+1)}) - \MLR(\bth,\bla^{{(k)}})\right),\, \bla^{(k)}\in\cG_{\bla}\right\}.
\end{eqnarray*}
The Ridge parameter $\bla$ is \textbf{not an hyperparameter} of our method; it is trained alongside the weights of the Neural Net architecture.

\bigskip
\paragraph{Choice of the number of iterations during the train.}
\begin{itemize}
    \item [$\bullet$] We fix the maximum number of iterations $\texttt{max}_{\iter}$ (depending on the value of $L$).
    \item [$\bullet$]  We fix the budget (\fixB = 5 min) and denote by $n_{\iter}$ the possible number of iterations during the allotted time \fixB. 
    \item [$\bullet$] We denote by $\iter$ the number of iterations that will actually be performed, $i.e.$
$$\iter= \min(\texttt{max}_{\iter},n_{\iter})$$
\end{itemize}

 \bigskip
  
\paragraph{Training \MLR{}-NN.}

We train the \FFNN{} with $\bs=\min(J,n)$ and we use Adam \cite{kingma2014adam} with default parameters except for the learning rate $\lr$ which depends on the number of layers $L$ (Table~\ref{tab:architectures1}). 
\medskip

\begin{mdframed}

$
\begin{array}{l}
    \textbf{Training}\\
      \quad \left|\begin{array}{llll}
          \textbf{Initialization}\\
                \quad \left| \begin{array}{ll}
                   \Set\,\bth \\
                     \Set\,\bla \\
                \end{array} \right.\\
    \textbf{Optimization}\\
    \quad \left| \begin{array}{ll}
         \While\,\, e < \iter \,\,\,\Do:\\
         \quad \left| \begin{array}{llllll}
           \bA^{0}-> \bx\in\R^{\bs\times d}\\
          \For\,\, \ell = 1 \cdots L-1:\\
               \quad \left|\begin{array}{l}
                           \bA^{\el} ->
                     \ReLu(\bA^{\el-1}W^{\el} +B^{\el})
                       \end{array} \right.\\
             \bH(\bth,\bla) -> \bA^{L-1}\left[(\bA^{L-1})^\top\bA^{L-1}+\bla \I_J\right]^{-1}{\bA^{L-1}}^\top\\
             \texttt{Compute }\, \MLR(\bth,\bla) \quad \text{or}\quad \CEMLR(\bth,\bla)
             \\
            \textbf{Backpropagate }\, (\bth,\bla) \textbf{ through }\, \MLR(\bth,\bla) \quad \text{or}\quad \CEMLR(\bth,\bla)\\
            e -> e+1\\
        \end{array} \right.\\
    \end{array} \right.\\
    \end{array} \right.\\
\end{array}
$
\end{mdframed}

We select a $validation$-set of size $n_{val}=20\%\, n$. We read the $\Rdeux$-score for each iteration on the $validation$-set and take the iteration with the best $\Rdeux$-score: 
$$\iter^*:=\arg\max\{\texttt{R}^2_k, \ k =1,\cdots,\iter\}.$$
Finally, $(\hbth,\hatla)$ will take its value at iteration $\iter^*$

\bigskip

\paragraph{Our final models.}
We propose several models with varying depth based on \FFNN{} trained with the \MLR{} loss. We also create ensemble models combining architectures of different depth.
Our models are:\\
$\bullet$  \MLRL: a simple \FFNN{} of depth $L$ ($1\leq L\leq 4$).\\
$\bullet$ \BMLRL: a bagging of 10 \FFNN{} of depth $L$ ($L=1$ or $L=2$).\\
$\bullet$ \BMLR{}: an ensemble of 20 \FFNN{} (the aggregation of \BMLRa{} and \BMLRb{} of depth $1$ and $2$ respectively).\\
$\bullet$ \BestMLR{}: the best prediction among 20 \NN{} in terms of the validation score.\\
$\bullet$ \TopfiveMLR{}: the aggregation of the top 5 among 20 \NN{} in terms of the validation score.

For the methods based on bagging \cite{breiman1996}, the final prediction is the mean of each \NN{} prediction.

\section{Construction of the Benchmark}


To produce this benchmark (Table~\ref{tab:datasets}), we aggregated 32 tabular datasets (16 in regression  and 16 in classification), from the UCI repository and Kaggle. 
For computational reasons, we have chosen to restrict the number of datasets but we performed more $train$/$test$ splitting in order to reduce the variance of our results. 
We curated the UCI repository and Kaggle through a set of rules ($e.g.$ discard empty or duplicate datasets, times series, missing target, non $i.i.d.$ samples, text format, $etc$.).
\begin{table}[H]
\caption{Benchmark datasets. \# Num. and \# Cat. denote the initial number of numerical and categorical features respectively. We denote by $d$ the number of features after the pre-processing and one-hot encoding.}
\label{tab:datasets}
\centering
\footnotesize
\begin{tabular}{|l|c|c|c|c|c|}
\hline
Description & Task & $n$ & $d$ & \# Num. &  \# Cat.  \\
\hline
\hline
Concrete Slump Test -2 & \reg  & $ 103$ & $8$ & $8$ & $0 $ \\
\hline
Concrete Slump Test -3 & \reg  & $ 103$ & $8$ & $8$ & $0 $ \\
\hline
Concrete Slump Test -1 & \reg  & $ 103$ & $8$ & $8$ & $0 $ \\
\hline
Servo & \reg  & $ 168$ & $24$ & $2$ & $4 $ \\
\hline
Computer Hardware & \reg  & $ 210$ & $7$ & $7$ & $0 $ \\
\hline
Yacht Hydrodynamics & \reg  & $ 308$ & $33$ & $5$ & $3 $ \\
\hline
QSAR aquatic toxicity & \reg  & $ 546$ & $34$ & $8$ & $3 $ \\
\hline
QSAR Bioconcentration classes  & \reg  & $ 779$ & $25$ & $8$ & $4 $ \\
\hline
QSAR fish toxicity & \reg  & $ 909$ & $18$ & $6$ & $2 $ \\
\hline
insurance & \reg  & $ 1338$ & $15$ & $3$ & $4 $ \\
\hline
Communities and Crime & \reg  & $ 1994$ & $108$ & $99$ & $2 $ \\
\hline
Abalone R & \reg  & $ 4178$ & $11$ & $7$ & $1 $ \\
\hline
squark automotive CLV training & \reg  & $ 8099$ & $77$ & $7$ & $16 $ \\
\hline
Seoul Bike Sharing Demand & \reg  & $ 8760$ & $15$ & $9$ & $3 $ \\
\hline
Electrical Grid Stability Simu & \reg  & $ 10000$ & $12$ & $12$ & $0 $ \\
\hline
blr real estate prices & \reg  & $ 13320$ & $2$ & $2$ & $0 $ \\
\hline
Cervical Cancer Behavior Risk & \Clf  & $ 72$ & $149$ & $19$ & $14 $ \\
\hline
Post-Operative Patient & \Clf  & $ 91$ & $32$ & $0$ & $8 $ \\
\hline
Breast Cancer Coimbra & \Clf  & $ 116$ & $9$ & $9$ & $0 $ \\
\hline
Heart failure clinical records & \Clf  & $ 299$ & $12$ & $7$ & $5 $ \\
\hline
Ionosphere & \Clf  & $ 352$ & $34$ & $32$ & $2 $ \\
\hline
Congressional Voting Records & \Clf  & $ 436$ & $64$ & $0$ & $16 $ \\
\hline
Cylinder Bands & \Clf  & $ 541$ & $111$ & $1$ & $19 $ \\
\hline
Credit Approval & \Clf  & $ 691$ & $42$ & $4$ & $8 $ \\
\hline
Tic-Tac-Toe Endgame & \Clf  & $ 959$ & $36$ & $0$ & $9 $ \\
\hline
QSAR biodegradation & \Clf  & $ 1056$ & $141$ & $41$ & $15 $ \\
\hline
Chess (King-Rook vs. King-Pawn & \Clf  & $ 3196$ & $102$ & $0$ & $36 $ \\
\hline
Mushroom & \Clf  & $ 8125$ & $125$ & $0$ & $21 $ \\
\hline
Electrical Grid Stability Simu & \Clf  & $ 10000$ & $12$ & $12$ & $0 $ \\
\hline
MAGIC Gamma Telescope & \Clf  & $ 19021$ & $10$ & $10$ & $0 $ \\
\hline
Adult & \Clf  & $ 32561$ & $34$ & $6$ & $5 $ \\
\hline
Internet Firewall Data & \Clf  & $ 65532$ & $11$ & $11$ & $0 $ \\
\hline
\end{tabular}
\end{table}

\subsection{Pre-processing}

To avoid biasing the benchmark towards specific methods and to get a result as general as possible, we only applied as little preprocessing as we could, without using any feature augmentation scheme. The goal is not to get the best possible performance on a given dataset but to compare the methods on equal ground.
We first removed uninformative features such as sample index. Categorical features with more than 12 modalities were discarded as learning embeddings is out of the scope of this benchmark. We also removed samples with missing target.

\paragraph{Target treatment.}
The target is centered and standardized via the function $\fR(\cdot)$. We remove the observation  when the value is missing.

\begin{mdframed}
$
\begin{array}{l}
\fR(Y)\\
\quad \left|\begin{array}{ll}
Y -> \text{float32}(Y)\\
\textbf{for}\,\,  i=1:n\\
\quad \left| \begin{array}{l}
\If\,\, Y_i==\NAN \\
\qquad\Remove(x_i,Y_i)\\
\end{array} \right.\\
Y -> \frac{Y-\overline{Y}}{\bar\sigma(Y)}\\
\end{array} \right.
\end{array}
$
\end{mdframed}

\bigskip

\paragraph{Features treatment.}
The imputation treatment is done during processing. For categorical  features, \NAN{} Data may be considered as a new class. For numerical features, we replace missing values by the mean. Set $n_j=\#\Set(X_j)$ the number of distinct values taken by the feature $X_j$, We proceed as follows :

\begin{itemize}
    \item[$\bullet$] When $n_j=1$, the feature $X_j$ is irrelevant, we remove it.
    \item[$\bullet$] When $n_j=2$ (including potentially \NAN{} class), we
    perform numerical encoding of binary categorical features.
    \item[$\bullet$]  Numerical features with less than $12$ distinct values are also treated as categorical features  ($2<n_j\leq 12$). We apply one-hot-encoding. 
    \item[$\bullet$] Finally, categorical features with $n_j> 12$ are removed.
   
\end{itemize}

\subsection{Compared methods}
\label{sec: ComparedMethod}


We ran the benchmark with \textbf{all the methods} (see Table~\ref{tab:methods}) \textbf{available in the scikit-learn library} for classification and regression (including \RF{} and \XGB) as well as the \GBDT{} methods. All methods were ran with the default hyperparameters. 

\begin{table}[H]
\caption{Main classes of methods.}
	\centering
	\footnotesize
\begin{tabular}{|l|l|}
\hline
Class      & \\
of Methods & Methods\\
\hline
\MLR{} (this paper)   & \MLRL, \BMLRL, \BMLR, \BestMLR, \TopfiveMLR  \\
\hline
\GBDT  & \XGB{} \cite{Breiman1997_arcing,Friedman2001,Friedman2002},  \Catboost{} \cite{Prokhorenkova2018}, \XGBoost{} \cite{guestrin2016}, \lightgbm{} \cite{Ke2017} \\
\hline
\RF  & \RF{} and \XRF{} \cite{breiman2001,barandiaran1998random}  \\
\hline               
\SVM &  \texttt{Lin-}\SVM{}, \SVM{}, $\nu$\texttt{-}\SVM{} \cite{Chihchung2011}\\
\hline
\RNN   & \texttt{Fast.ai} \cite{Howard2020}, \MLP \cite{Hinton89connectionistlearning}\\
\hline
\GLM   & \texttt{OLS}, \Enet{} \cite{Zou05}, \Ridge{} \cite{Hoerl1970}, \LAS{} \cite{tibshirani1996}, \texttt{Logistic regression} \cite{cox1958}\\
\hline
\MARS & \MARS{} \cite{Friedman1991}\\
\hline
\TREE& \texttt{CART}, \texttt{XCART} \cite{Breiman1984,gey2005model,Klusowski2020sparse}\\
\hline
\Baseline & Reg: \texttt{Intercept}$\,|\,$ Classif: \texttt{Class probabilities}\\
\hline
\end{tabular}
\end{table}

\section{\MLR{} Parameters Analysis} 

In this section we study the behavior of the \MLR{} method and the impact of its key components through extensive evaluation on three datasets, \UCIa, \UCIb{} and \UCIc{}, for which $(n,d)$ are equal to $(103,8)$, $(546,34)$ and $(8760,15)$ respectively. We repeated each experiment over 100 random $train$/$test$ splits.

\subsection{Impact of the MLR components.}

In this section, we study the impact of the different components in the \MLR{} approach on the
the $\Rdeux$-score on the $test$ and $validation$ sets, computation time, the  convergence of the method (\Iter) and the initialization of the Ridge parameter $\bla_{init}$. To study the impact of each specific parameter, we set the other ones equal to their default values in Table~\ref{tab:architectures1}. Note that for the following study, we chose a batch size $\bs=\min(n,2^{14})$, unlike in our main experiments where we took $\bs=\min(n,2^{10})$ due to time constraints.

Note also that due to access failure to \VM, computation time was sometimes obtained on a less powerful configuration in Tables \ref{tab:StructDithering}, \ref{tab:Permutation}, \ref{tab:lambdaInit} and \ref{tab:Dithering}. We marked by an asterisk $\boldsymbol{\ast}$ any computation time obtained on the \Portable{} configuration.

\paragraph{Structured Dithering.} 
Recall that we added Structured noise $(\I_n-\bH )\xi$ to the target $\bY$ with $\xi\sim \cN(0,\sigma^2\I)$.
Table~\ref{tab:StructDithering} reveals the impact of the structured dithering parameter $\sigma$. Default value ($\sigma = 1$) yields consistently good generalization performance. Of course, it is always possible to tune this hyperparameter around value $1$ for potential improvement of the generalization performances. Higher values of $\sigma$ lead to a significant degradation of $\Rdeux$-score as it caused the method to diverge. In our experiments, $\sigma$ was not an hyperparameter as it was always set equal to $1$. Moreover, adding structured dithering has no impact on the value of $\bla_{init}$ or computational time.

\begin{table}[H]
\caption{Structured dithering dependence.}
\label{tab:StructDithering}
	\centering
\begin{tabular}{|c|c|c|c|c|c|c|}
\hline
 \UCIa &$\sigma$ &  $\Rdeux$ & Time & \iter & $\Rdeux_{val}$ & $\bla_{init}$  \\
\hline
\hline
&0  & $ 0.321$ & $30.356$ & $60.650$ & $0.479$ & $219.345$ \\
\hline
&0.2  & $ 0.338$ & $30.424$ & $79.830$ & $0.496$ & $219.345 $ \\
\hline
&\textbf{1} & $ 0.357$ & $30.423$ & $99.570$ & $0.515$ & $219.345 $ \\
\hline
&2 & $ 0.089$ & $1.312$ & $0.250$ & $0.137$ & $219.345 $ \\
\hline
&3 & $ 0.068$ & $1.257$ & $0.000$ & $0.116$ & $219.345 $ \\
\hline
\hline
\UCIb & $\sigma$ &  $\Rdeux$  & Time & \iter & $\Rdeux_{val}$  & $\bla_{init}$  \\
\hline
\hline
&0  & $ 0.463$ & $32.250$ & $11.200$ & $0.511$ & $774.264 $ \\
\hline
&0.2  & $ 0.463$ & $32.408$ & $14.550$ & $0.514$ & $774.264 $ \\
\hline
&\textbf{1} & $ 0.460$ & $32.281$ & $46.750$ & $0.525$ & $774.264 $ \\
\hline
&2 & $ 0.220$ & $1.276$ & $0.020$ & $0.226$ & $774.264 $ \\
\hline
&3 & $ 0.216$ & $1.288$ & $0.000$ & $0.223$ & $774.264 $ \\
\hline
\hline
 \UCIc &$\sigma$  &  $\Rdeux$  & Time & \iter & $\Rdeux_{val}$  & $\bla_{init}$  \\
\hline
\hline
&0  & $ 0.863$ & $89.425$ & $181.300$ & $0.864$ & $10000.001 $ \\
\hline
&0.2  & $ 0.863$ & $90.206$ & $188.520$ & $0.864$ & $10000.001 $ \\
\hline
&\textbf{1} & $ 0.855$ & $89.968$ & $191.920$ & $0.857$ & $10000.001 $ \\
\hline
&2  & $ 0.364$ & $1.876$ & $0.000$ & $0.363$ & $10000.001 $ \\
\hline
&3 & $ 0.364$ & $1.891$ & $0.000$ & $0.363$ & $10000.001 $ \\
\hline
\end{tabular}
\end{table}

\paragraph{Permutations.}
We studied the impact of the randomness aspect of the \MLR{} loss. We compared different sets of permutations drawn at random. The choice of the seed has little impact on the value of the \MLR{} loss as soon as $T\geq 2^2$. Table~\ref{tab:Permutation} reveals a significant jump in $\Rdeux$-score on the test going from $T=0$ to $T=1$ permutation. Then, increasing the value of $T>1$ may sometimes slightly improve $\Rdeux$-score.
Meanwhile, a larger number of permutations has a direct negative impact on runtime per iteration and VRAM footprint. Past a certain threshold $2^8$, GPU parallelization no longer prevents the linear dependency on $T$. We escape any trade-off by picking $T=2^4$ permutations in all our experiments. This value is large enough for the \MLR{} loss to converge (with regards to $T$), yet still leveraging GPU parallelization.

\begin{table}[H]
	\caption{Permutation dependence. $\boldsymbol{\ast}$: computation time was obtained with a \Portable.}
		\label{tab:Permutation}
	\centering
\begin{tabular}{|c|c|c|c|c|c|c|}
\hline
\UCIa & $T$  &  $\Rdeux$ & Time & \iter & $\Rdeux_{val}$ & $\bla_{init}$  \\
\hline
\hline
 &  0  & $ 0.252$ & $3.184$ & $61.050$ & $0.401$ & $31.831 $ \\
\hline
 &  1  & $ 0.331$ & $3.357$ & $110.040$ & $0.459$ & $285.238 $ \\
\hline
 &  2  & $ 0.338$ & $3.359$ & $109.960$ & $0.468$ & $215.370 $ \\
\hline
 &  $2^{2}$  & $ 0.343$ & $3.358$ & $109.370$ & $0.473$ & $219.345 $ \\
\hline
 &  $\boldsymbol{2^{4}}$  & $ 0.347$ & $4.012^{\textbf{*}}$ & $116.190$ & $0.484$ & $216.235 $ \\
\hline
 &  $2^{8}$  & $ 0.351$ & $3.371$ & $117.160$ & $0.494$ & $219.345 $ \\
\hline
 &  $2^{10}$  & $ 0.349$ & $3.433$ & $117.650$ & $0.495$ & $219.345 $ \\
\hline
\hline
\UCIb & $T$  &  $\Rdeux$ & Time & \iter & $\Rdeux_{val}$ & $\bla_{init}$  \\
\hline
\hline
 &  0.0  & $ 0.460$ & $3.253$ & $46.770$ & $0.509$ & $774.264 $ \\
\hline
 &  1  & $ 0.461$ & $3.452$ & $62.020$ & $0.518$ & $774.264 $ \\
\hline
 &  2  & $ 0.466$ & $3.461$ & $60.040$ & $0.518$ & $774.264 $ \\
\hline
 &  $2^{2}$  & $ 0.469$ & $3.462$ & $60.720$ & $0.521$ & $774.264 $ \\
\hline
 &  $\boldsymbol{2^{4}}$  & $ 0.473$ & $6.172^{\textbf{*}}$ & $72.800$ & $0.527$ & $774.264 $ \\
\hline
 &  $2^{8}$  & $ 0.477$ & $3.496$ & $81.900$ & $0.532$ & $774.264 $ \\
\hline
 &  $2^{10}$  & $ 0.480$ & $3.551$ & $81.530$ & $0.532$ & $774.264 $ \\
\hline
\hline
\UCIc & $T$ &  $\Rdeux$ & Time & \iter & $\Rdeux_{val}$ & $\bla_{init}$  \\
\hline
\hline
 &  0  & $ 0.817$ & $8.251$ & $197.830$ & $0.817$ & $10000.001 $ \\
\hline
 &  1  & $ 0.813$ & $8.606$ & $197.860$ & $0.813$ & $10000.001 $ \\
\hline
 &  2  & $ 0.813$ & $8.654$ & $197.400$ & $0.814$ & $10000.001 $ \\
\hline
 &  $2^{2}$  & $ 0.813$ & $8.645$ & $197.780$ & $0.814$ & $10000.001 $ \\
\hline
 &  $\boldsymbol{2^{4}}$  & $ 0.814$ & $30.654^{\textbf{*}}$ & $197.100$ & $0.814$ & $10000.001 $ \\
\hline
 &  $2^{8}$  & $ 0.813$ & $10.391$ & $197.230$ & $0.814$ & $10000.001 $ \\
\hline
 &  $2^{10}$  & $ 0.814$ & $17.330$ & $197.070$ & $0.814$ & $10000.001 $ \\
\hline
\end{tabular}
\end{table}

\paragraph{Initialization of Ridge parameter $\bla_{init}$.}
Recall that Ridge regularization is the essential component of the \MLR{} method as it provides a closed form representation of the last hidden layer on which we can conveniently apply the follow-up steps: structured dithering and random permutations. Contrary to $T$ and the dither parameter $\sigma$, the choice of the appropriate initial value of $\bla$ is very impactful and depends on both network architecture and dataset characteristics as shown in Table~\ref{tab:lambdaInit}.

When we compare the value $\bla_{init}$ given by our heuristic (in bold) with the other values chosen in Table~\ref{tab:lambdaInit}, we observe that our heuristic is quite effective, as in average on the 3 datasets, it is always within $3\%$ of the best value in the grid of Table \ref{tab:lambdaInit} in term of $\Rdeux$-score on the $test$. As we can see for the \UCIb{} dataset, the optimal value was not within the bounds of the grid $\cG_{\bla}$ we chose. Using a larger grid with a bigger granularity would improve the results.

Despite access failure to \VM{} for one specific value of $\bla_{init}$, our main experiments reveal a small runtime overcost for the initialization step, 
mostly because all steps including the matrix inversion need to be performed only once and do not require computing the derivation graph. We favored a small simple grid 
$\cG_{\bla}=\left\lbrace \bla^{(k)} = 10^{-1}\times10^{5 \times k / 11}\;: \;   k = 0, \cdots, 11 \right\rbrace$ to select $\bla_{init}$. This grid was designed to work well on small size datasets. Of course, it is possible to refine this grid with respect to the dataset size and architecture at hand to achieved even higher generalization performance.
Another possible approach could be to tune $\bla_{init}$ on the $validation$ set. Indeed, we observe in Table \ref{tab:lambdaInit} that the optimal value of $\bla_{init}$ on the $test$ seems to be close to that obtained on the $validation$ set.

\begin{table}[H]
	\caption{Dependence on $\bla_{init}$. $\boldsymbol{\ast}$: computation time was obtained with a \Portable.}
		\label{tab:lambdaInit}
	\centering
\begin{tabular}{|c|c|c|c|c|c|}
\hline
\UCIa &$\bla_{init}$ & $\Rdeux$ & Time & \iter & $\Rdeux_{val}$  \\
\hline
\hline
 &  0  & $ -0.110$ & $0.180$ & $7.790$ & $-0.020 $ \\
\hline
&$10^{-3}$  & $ -0.444$ & $2.078$ & $90.270$ & $0.265 $ \\
\hline
&$10^{-1}$  & $ 0.097$ & $2.083$ & $70.310$ & $0.254 $ \\
\hline
&$10$  & $ 0.320$ & $2.070$ & $116.630$ & $0.466 $ \\
\hline
 &  $\boldsymbol{216.235}$  & $ 0.347$ & $2.902^{\textbf{*}}$ & $116.190$ & $0.484 $ \\
\hline
&$10^{3}$  & $ 0.359$ & $2.087$ & $125.020$ & $0.480 $ \\
\hline
&$10^{5}$  & $ 0.334$ & $2.103$ & $152.460$ & $0.428 $ \\
\hline
&$10^{7}$  & $ 0.263$ & $2.104$ & $188.630$ & $0.339 $ \\
\hline
&$10^{9}$  & $ -0.050$ & $2.089$ & $197.890$ & $-0.009 $ \\
\hline
\hline
 \UCIb  &$\bla_{init}$ & $\Rdeux$ & Time & \iter & $\Rdeux_{val}$  \\
\hline
\hline
 &  0  & $ -0.276$ & $0.014$ & $0.010$ & $-0.244 $ \\
\hline
&$10^{-3}$  & $ -33.053$ & $0.133$ & $2.510$ & $-9.371 $ \\
\hline
&$10^{-1}$  & $ -3.768$ & $2.137$ & $36.770$ & $-0.151 $ \\
\hline
&$10$  & $ 0.422$ & $2.086$ & $9.530$ & $0.477 $ \\
\hline
 &  $\boldsymbol{774.263}$  & $ 0.473$ & $3.426^{\textbf{*}}$ & $72.800$ & $0.527 $ \\
\hline
&$10^{3}$  & $ 0.477$ & $2.094$ & $73.530$ & $0.529 $ \\
\hline
&$10^{5}$  & $ 0.486$ & $2.088$ & $132.420$ & $0.522 $ \\
\hline
&$10^{7}$  & $ 0.477$ & $2.088$ & $191.320$ & $0.488 $ \\
\hline
&$10^{9}$  & $ 0.273$ & $2.086$ & $200.000$ & $0.287 $ \\
\hline
\hline
 \UCIc  &$\bla_{init}$ & $\Rdeux$ & Time & \iter & $\Rdeux_{val}$  \\
\hline
\hline
 &  0.0  & $ -0.091$ & $0.052$ & $0.010$ & $-0.088 $ \\
\hline
&$10^{-3}$  & $ 0.761$ & $5.042$ & $97.970$ & $0.775 $ \\
\hline
&$10^{-1}$  & $ 0.795$ & $5.009$ & $66.370$ & $0.807 $ \\
\hline
&$10$  & $ 0.844$ & $4.989$ & $161.160$ & $0.847 $ \\
\hline
&$10^{3}$  & $ 0.843$ & $4.974$ & $194.550$ & $0.844 $ \\
\hline
 &  $\boldsymbol{10^4}$ & $ 0.814$ & $19.208^{\textbf{*}}$ & $197.100$ & $0.814 $ \\
\hline
&$10^{5}$  & $ 0.774$ & $4.966$ & $197.510$ & $0.775 $ \\
\hline
&$10^{7}$  & $ 0.711$ & $4.956$ & $198.600$ & $0.710 $ \\
\hline
&$10^{9}$  & $ 0.614$ & $4.942$ & $198.830$ & $0.613 $ \\
\hline
\end{tabular}
\end{table}

\paragraph{Ablation study.}
We ran our ablation study (Table \ref{tab:ablation}) in the regression setting on the same 3 datasets (\UCIa, \UCIb, \UCIc). We repeated each experiment over 100 random $train$/$test$ splits. All the results presented here correspond to the architecture of \MLRb{} and \BMLRb{} with hyperparameters fixed as in Table~\ref{tab:architectures1}.

A standard \RNN2 (\FFNN{} with $2$ wide layers $J=2^{10}$) cannot be trained efficiently on  small datasets as the \FFNN{} instantly memorizes the entire dataset. This cannot be alleviated through bagging at all. Note also its lower overall performance on the complete benchmark.

Applying Ridge on the last hidden layer allows an extremely overparametrized \FFNN{} to learn but its generalization performance is still far behind the gold standard \RF{}. However, when using bagging with ten such models, we reach very competitive results, underlying the potential of the \MLR{} approach.

The random permutations component gives a larger improvement than Structured Dithering. However, when using both ingredients together, a single \NN{} can reach or even outperform the gold-standard methods on most datasets. Furthermore, the improvement yielded by using bagging ($0.062$) is still of the same order of magnitude as the one we got when we applied permutations on top of Ridge to the \FFNN{} ($0.043$). This means these two ingredients (permutations and Structure Dithering) are not just simple variance reduction techniques but actually generate more sophisticated models.

\begin{table}[H]
\caption{Ablation Study in Regression.}
	\centering
	\footnotesize
\begin{tabular}{|l||c|c|}
\hline
Step  &  Mean $\Rdeux$  & Bagging $\Rdeux$ \\
\hline
\hline
\RNN2  & $ -0.081  \pm  0.173 $ & $ -0.046  \pm  0.169 $ \\
\hline
\FFNN{}+ Ridge  & $ 0.321  \pm  0.081 $ & $ 0.394  \pm  0.052 $ \\
\hline
\FFNN{}+ Ridge + Struct. Dithering  & $ 0.323  \pm  0.075 $ & $ 0.400  \pm  0.048 $ \\
\hline
\FFNN{}+ Ridge + Permut. & $ 0.364  \pm  0.050 $ & $ 0.432  \pm  0.035 $ \\
\hline
\MLR{}   & $ 0.371  \pm  0.024 $ & $ 0.433  \pm  0.000 $ \\
\hline
\end{tabular}
\end{table}

\subsection{Other hyperparameters.} 

The impact of the other hyperparameters on the \MLR{} method is discussed below.

\paragraph{\Dither.}
At each iteration, we draw and add i.i.d. gaussian noise $\cN(0,\tilde{\sigma}^2\I)$ on the target $\bY$ in the regression setting. In Table~\ref{tab:Dithering}, we see that adding a small amount of noise improves performances. We performed our main experiments with $\tilde{\sigma} = 0.03$ as this value works well with standard \FFNN{}. But here again, we may improve generalization performance by considering $\tilde{\sigma}$ as an hyperparameter to be tuned. Rather unsurprisingly, applying dithering has no impact on runtime per iteration or on the value of $\bla_{init}$.

\begin{table}[H]
\caption{Dithering dependence : label noise scale. $\boldsymbol{\ast}$: computation time was obtained with a \Portable.}
\label{tab:Dithering}
	\centering
\begin{tabular}{|c|c|c|c|c|c|c|}
\hline
\UCIa & $\tilde{\sigma}$  &  $\Rdeux$ & Time & \iter & $\Rdeux_{val}$ & $\bla_{init}$  \\
\hline
\hline
 &  0  & $ 0.347$ & $3.104$ & $116.490$ & $0.483$ & $220.900 $ \\
\hline
 &  0.01  & $ 0.351$ & $3.110$ & $114.560$ & $0.486$ & $220.900 $ \\
\hline
 &  $\boldsymbol{0.03}$  & $ 0.347$ & $5.560^{\textbf{*}}$ & $116.190$ & $0.484$ & $216.235 $ \\
\hline
 &  0.1  & $ 0.354$ & $3.111$ & $113.720$ & $0.490$ & $215.104 $ \\
\hline
 &  0.3  & $ 0.353$ & $3.108$ & $119.300$ & $0.502$ & $216.367 $ \\
\hline
\UCIb  & $\tilde{\sigma}$ & $\Rdeux$ & Time & \iter & $\Rdeux_{val}$ & $\bla_{init}$  \\
\hline
\hline
 &  0 & $ 0.475$ & $3.250$ & $76.830$ & $0.527$ & $774.264 $ \\
\hline
 &  0.01  & $ 0.474$ & $3.258$ & $68.680$ & $0.527$ & $774.264 $ \\
\hline
 &  $\boldsymbol{0.03}$  & $ 0.473$ & $9.709^{\textbf{*}}$ & $72.800$ & $0.527$ & $774.264 $ \\
\hline
 &  0.1  & $ 0.474$ & $3.258$ & $70.860$ & $0.528$ & $774.264 $ \\
\hline
 &  0.3  & $ 0.474$ & $3.258$ & $68.620$ & $0.532$ & $774.264 $ \\
\hline
\UCIc  & $\tilde{\sigma}$ & $\Rdeux$ & Time & \iter & $\Rdeux_{val}$ & $\bla_{init}$  \\
\hline
\hline
 &  0  & $ 0.813$ & $8.554$ & $197.430$ & $0.814$ & $10000.001 $ \\
\hline
 &  0.01  & $ 0.814$ & $8.561$ & $197.240$ & $0.814$ & $10000.001 $ \\
\hline
 &  $\boldsymbol{0.03}$  & $ 0.814$ & $29.283^{\textbf{*}}$ & $197.100$ & $0.814$ & $10000.001 $ \\
\hline
 &  0.1  & $ 0.813$ & $8.561$ & $196.720$ & $0.814$ & $10000.001 $ \\
\hline
 &  0.3  & $ 0.812$ & $8.567$ & $196.220$ & $0.813$ & $10000.001 $ \\
\hline
\end{tabular}
\end{table}

\paragraph{Width.}
Most notably, Table~\ref{tab:Width} reveals that wide architectures (large $J$) usually provide better generalization performance. We recall that for standard \RNN{} trained without \MLR, wider architectures are more prone to overfitting. Table \ref{tab:Width} also reveals that larger architectures work better for bigger datasets like \UCIc. For small datasets, $J=2^{10}$ provides good generalization performance for smaller runtime. When the width parameter exceeds GPU memory, parallelization is lost and we observe a dramatic increase in computational time.

\begin{table}[H]
\caption{Width dependence.}
\label{tab:Width}
	\centering
\begin{tabular}{|c|c|c|c|c|c|c|}
\hline
\UCIa  &$J$ & $\Rdeux$ & Time & \iter & $\Rdeux_{val}$ &
$\bla_{init}$  \\
\hline
\hline
&$2^{4}$  & $ 0.184$ & $0.705$ & $162.120$ & $0.284$ & $210.307 $ \\
\hline
&$2^{6}$ & $ 0.276$ & $0.751$ & $160.030$ & $0.364$ & $211.555 $ \\
\hline
&$2^{8}$  & $ 0.325$ & $0.905$ & $135.400$ & $0.431$ & $205.351 $ \\
\hline
&$\boldsymbol{2^{10}}$  & $ 0.344$ & $2.201$ & $113.610$ & $0.484$ & $222.455 $ \\
\hline
&$2^{12}$  & $ 0.322$ & $15.796$ & $94.180$ & $0.503$ & $220.900 $ \\
\hline
\hline
 \UCIb   &$J$ & $\Rdeux$ & Time & \iter & $\Rdeux_{val}$ &
$\bla_{init}$  \\
\hline
\hline
&$2^{4}$  & $ 0.367$ & $0.724$ & $184.540$ & $0.379$ & $678.097 $ \\
\hline
&$2^{6}$ & $ 0.442$ & $0.743$ & $157.840$ & $0.471$ & $628.464 $ \\
\hline
&$2^{8}$  & $ 0.467$ & $0.907$ & $115.510$ & $0.512$ & $774.264 $ \\
\hline
&$\boldsymbol{2^{10}}$  & $ 0.470$ & $2.188$ & $71.790$ & $0.527$ & $774.264 $ \\
\hline
&$2^{12}$  & $ 0.460$ & $16.987$ & $37.210$ & $0.524$ & $774.264 $ \\
\hline
\hline
 \UCIc  &$J$ & $\Rdeux$ & Time & \iter & $\Rdeux_{val}$ &
$\bla_{init}$  \\
\hline
\hline
&$2^{4}$  & $ 0.622$ & $1.008$ & $200.000$ & $0.620$ & $9350.431 $ \\
\hline
&$2^{6}$ & $ 0.714$ & $1.134$ & $200.000$ & $0.713$ & $9927.827 $ \\
\hline
&$2^{8}$  & $ 0.773$ & $1.955$ & $199.880$ & $0.773$ & $10000.001 $ \\
\hline
&$\boldsymbol{2^{10}}$  & $ 0.825$ & $7.062$ & $198.240$ & $0.825$ & $10000.001 $ \\
\hline
&$2^{12}$  & $ 0.856$ & $54.121$ & $193.270$ & $0.857$ & $10000.001 $ \\
\hline
\end{tabular}
\end{table}

\paragraph{Batch size.} We added the \UCId{} of size $(n,d)=(43824,33)$ in this experiment in order to measure the impact of batch-size on a larger dataset
but this dataset was not included in the benchmark.

In view of Table~\ref{tab:Batchsize}, our recommendation is very simple: "\textbf{\textit{As big as possible !}}". For small datasets this means using the entire train-set at each iteration, while GPU memory constraints rule out going beyond $2^{14}$ for large datasets.

\begin{table}[H]
	\centering
	\resizebox{\columnwidth}{!}{
\begin{tabular}{|c|c|c|c|c|c|c|}
\hline
 \UCIa &$\bs$  & $\Rdeux$ & Time & \iter & $\Rdeux_{val}$ & $\bla_{init}$  \\
\hline
\hline
&1  & $ -0.122$ & $4.375$ & $32.596$ & $0.014$ & $38.452 $ \\
\hline
&$2^{4}$  & $ 0.334$ & $5.194$ & $129.673$ & $0.520$ & $82.567 $ \\
\hline
&$2^{5}$  & $ 0.349$ & $5.194$ & $107.269$ & $0.517$ & $110.214 $ \\
\hline
&$2^{6}$  & $ 0.393$ & $5.352$ & $115.115$ & $0.500$ & $246.869 $ \\
\hline
&$\boldsymbol{\min(n,2^{14})=103}$    & $ 0.401$ & $5.238$ & $114.385$ & $0.499$ & $237.899 $ \\
\hline
\hline
\UCIb &$\bs$   & $\Rdeux$ & Time & \iter & $\Rdeux_{val}$ & $\bla_{init}$  \\
\hline
\hline
&1  & $ -0.014$ & $4.658$ & $38.020$ & $0.003$ & $290.923 $ \\
\hline
&$2^{4}$  & $ 0.415$ & $5.046$ & $148.680$ & $0.490$ & $158.198 $ \\
\hline
&$2^{5}$  & $ 0.459$ & $5.180$ & $141.260$ & $0.527$ & $204.647 $ \\
\hline
&$2^{6}$  & $ 0.474$ & $5.216$ & $128.820$ & $0.545$ & $253.497 $ \\
\hline
&$2^{7}$  & $ 0.477$ & $5.277$ & $103.270$ & $0.540$ & $388.678 $ \\
\hline
&$2^{8}$  & $ 0.478$ & $5.254$ & $97.010$ & $0.535$ & $774.264 $ \\
\hline
&$\boldsymbol{\min(n,2^{14})=546}$  & $ 0.475$ & $5.301$ & $72.470$ & $0.528$ & $774.264 $ \\
\hline
\hline
\UCIc &$\bs$   & $\Rdeux$ & Time & \iter & $\Rdeux_{val}$ & $\bla_{init}$  \\
\hline

\hline
&1  & $ 0.013$ & $4.536$ & $15.790$ & $0.013$ & $89.257 $ \\
\hline
&$2^{4}$  & $ 0.640$ & $5.317$ & $168.790$ & $0.642$ & $107.543 $ \\
\hline
&$2^{5}$  & $ 0.673$ & $5.375$ & $176.200$ & $0.675$ & $171.905 $ \\
\hline
&$2^{6}$  & $ 0.703$ & $5.360$ & $186.940$ & $0.705$ & $212.625 $ \\
\hline
&$2^{7}$  & $ 0.729$ & $5.413$ & $188.540$ & $0.730$ & $537.572 $ \\
\hline
&$2^{8}$  & $ 0.750$ & $5.447$ & $189.770$ & $0.751$ & $774.264 $ \\
\hline
&$2^{9}$  & $ 0.763$ & $5.460$ & $191.490$ & $0.765$ & $2641.979 $ \\
\hline
&$2^{10}$  & $ 0.785$ & $5.781$ & $192.900$ & $0.786$ & $2782.560 $ \\
\hline
&$2^{11}$  & $ 0.790$ & $6.908$ & $195.150$ & $0.791$ & $10000.001 $ \\
\hline
&$2^{12}$  & $ 0.813$ & $9.842$ & $196.930$ & $0.814$ & $10000.001 $ \\
\hline
&$\boldsymbol{\min(n,2^{14})=8760}$   & $ 0.824$ & $13.547$ & $197.800$ & $0.825$ & $10000.001 $ \\
\hline
\hline
\UCId &$\bs$  & $\Rdeux$ & Time & \iter & $\Rdeux_{val}$ &
$\bla_{init}$  \\
\hline

\hline
&1  & $ -0.003$ & $4.826$ & $65.470$ & $-0.003$ & $185.268 $ \\
\hline
&$2^{4}$  & $ 0.019$ & $5.344$ & $121.380$ & $0.021$ & $628.786 $ \\
\hline
&$2^{5}$  & $ 0.050$ & $5.629$ & $157.860$ & $0.052$ & $563.540 $ \\
\hline
&$2^{6}$  & $ 0.109$ & $5.659$ & $176.680$ & $0.110$ & $770.180 $ \\
\hline
&$2^{7}$  & $ 0.148$ & $5.646$ & $171.810$ & $0.149$ & $705.241 $ \\
\hline
&$2^{8}$  & $ 0.188$ & $5.678$ & $179.850$ & $0.190$ & $988.136 $ \\
\hline
&$2^{9}$  & $ 0.209$ & $5.771$ & $185.860$ & $0.211$ & $911.264 $ \\
\hline
&$2^{10}$  & $ 0.231$ & $6.165$ & $190.310$ & $0.233$ & $1001.640 $ \\
\hline
&$2^{11}$  & $ 0.254$ & $7.251$ & $192.960$ & $0.255$ & $1245.079 $ \\
\hline
&$2^{12}$  & $ 0.278$ & $10.199$ & $194.410$ & $0.279$ & $1376.753 $ \\
\hline
&$2^{13}$  & $ 0.298$ & $20.653$ & $195.740$ & $0.298$ & $2782.560 $ \\
\hline
&$\boldsymbol{\min(n,2^{14})=43824}$  & $ 0.316$ & $56.976$ & $193.520$ & $0.317$ & $3793.002 $ \\
\hline
\end{tabular}
	\caption{
Batch size dependence
		\label{tab:Batchsize}
}
}
\end{table}

\paragraph{Depth.} 

As we can see in Table~\ref{tab:Depth}, the optimal choice of the depth parameter seems to be data-dependent and significantly impacts the $\Rdeux$-score. This motivated the introduction of the bagging \MLR{} models that we described in the main paper. 

We consider only architectures of depth $L\in\cG_L=\{1,\,2,\,3,\, 4\}$ which reached state of the art results nonetheless. Going deeper is outside of the scope we set for this study, since it would probably require more careful and manual tuning of the hyperparameters on each dataset.

\begin{table}[H]
\footnotesize{
\caption{Depth dependence. Mean and standard deviation of $\Rdeux$-score over $100$ seeds.}
\label{tab:Depth}
	\centering
	\resizebox{\columnwidth}{!}{
\begin{tabular}{|l|c|c|c|c|c|c|}
\hline
Dataset & $n$ & $d$ & $\MLRa$& $\MLRb$& $\MLRc$& $\MLRd$\\
\hline
{Concrete Slump Test -1} & 103 & 8 & $0.940 \pm 0.029$ & $\boldsymbol{0.954 \pm 0.018}$ & $\boldsymbol{0.954 \pm 0.025}$ & $0.935 \pm 0.032$ \\
\hline
{Concrete Slump Test -3} & 103 & 8 & $\boldsymbol{0.399 \pm 0.132}$ & $0.313 \pm 0.171$ & $0.274 \pm 0.210$ & $0.226 \pm 0.149$ \\
\hline
{Concrete Slump Test -2} & 103 & 8 & $0.455 \pm 0.133$ & $0.453 \pm 0.159$ & $\boldsymbol{0.505 \pm 0.171}$ & $0.425 \pm 0.245$ \\
\hline
{Servo} & 168 & 24 & $0.836 \pm 0.031$ & $0.839 \pm 0.046$ & $\boldsymbol{0.854 \pm 0.043}$ & $0.842 \pm 0.049$ \\
\hline
{Computer Hardware} & 210 & 7 & $0.981 \pm 0.008$ & $0.984 \pm 0.008$ & $\boldsymbol{0.985 \pm 0.008}$ & $\boldsymbol{0.985 \pm 0.007}$ \\
\hline
{Yacht Hydrodynamics} & 308 & 33 & $0.952 \pm 0.021$ & $0.962 \pm 0.020$ & $\boldsymbol{0.965 \pm 0.020}$ & $0.960 \pm 0.021$ \\
\hline
{QSAR aquatic toxicity} & 546 & 34 & $0.448 \pm 0.081$ & $0.459 \pm 0.071$ & $0.458 \pm 0.090$ & $\boldsymbol{0.470 \pm 0.087}$ \\
\hline
{QSAR Bioconcentration classes}  & 779 & 25 & $\boldsymbol{0.672 \pm 0.042}$ & $0.668 \pm 0.049$ & $0.666 \pm 0.051$ & $0.670 \pm 0.051$ \\
\hline
{QSAR fish toxicity} & 909 & 18 & $\boldsymbol{0.590 \pm 0.043}$ & $0.586 \pm 0.037$ & $0.582 \pm 0.043$ & $0.579 \pm 0.046$ \\
\hline
{insurance} & 1338 & 15 & $\boldsymbol{0.839 \pm 0.024}$ & $0.837 \pm 0.026$ & $0.833 \pm 0.033$ & $0.832 \pm 0.028$ \\
\hline
{Communities and Crime} & 1994 & 108 & $0.679 \pm 0.031$ & $0.677 \pm 0.029$ & $\boldsymbol{0.680 \pm 0.030}$ & $\boldsymbol{0.680 \pm 0.027}$ \\
\hline
{Abalone R} & 4178 & 11 & $\boldsymbol{0.566 \pm 0.023}$ & $0.543 \pm 0.078$ & $0.523 \pm 0.163$ & $0.538 \pm 0.078$ \\
\hline
{squark automotive CLV training} & 8099 & 77 & $\boldsymbol{0.891 \pm 0.006}$ & $0.890 \pm 0.006$ & $0.889 \pm 0.006$ & $0.883 \pm 0.007$ \\
\hline
{Seoul Bike Sharing Demand} & 8760 & 15 & $0.850 \pm 0.009$ & $0.878 \pm 0.008$ & $\boldsymbol{0.901 \pm 0.008}$ & $0.893 \pm 0.008$ \\
\hline
{Electrical Grid Stability Simu} & 10000 & 12 & $0.937 \pm 0.003$ & $0.958 \pm 0.002$ & $\boldsymbol{0.963 \pm 0.002}$ & $0.955 \pm 0.002$ \\
\hline
{blr real estate prices} & 13320 & 2 & $0.514 \pm 0.012$ & $\boldsymbol{0.522 \pm 0.012}$ & $\boldsymbol{0.522 \pm 0.013}$ & ${0.521 \pm 0.013}$ \\
\hline
\end{tabular}}
}
\end{table}

\paragraph{Learning rate.} We used ADAM with default parameters except for the learning rate. Indeed, since the width and batch size we picked were outside of the usual ranges, we had to adjust the learning rate accordingly (Table~\ref{tab:architectures1}). We did not attempt to use another optimizer as ADAM worked well.

\paragraph{Scalability.} 

The main limitation is the size of the GPU VRAM with a current maximum of $32$G on the best available configuration.  We conducted these experiments on devices with either $8$ or $11$G$-$VRAM.
 
Recall that the cost for the inversion of a 
 $J\times J$ matrix is linear on a GPU thanks to parallelization whereas it is quadratic on a CPU.
 
 The runtime per iteration is almost constant since it depends mostly on width, depth, batch-size and number of permutations which are either fix or bounded (for batch-size).

\end{document}